\documentclass[11pt]{article}

\usepackage[preprint]{acl}

\usepackage{times}
\usepackage{latexsym}

\usepackage[T1]{fontenc}

\usepackage[utf8]{inputenc}

\usepackage{microtype}

\usepackage{inconsolata}

\usepackage{graphicx}

\usepackage{amsmath}
\usepackage{amsthm}
\usepackage{booktabs}
\usepackage{algorithm}
\usepackage{algorithmic}
\usepackage[switch]{lineno}
\usepackage{amsfonts}
\usepackage{tcolorbox}
\usepackage{multirow}
\usepackage[table]{xcolor}
\usepackage{array}
\newcommand{\PreserveBackslash}[1]{\let\temp=\\#1\let\\=\temp}
\newcolumntype{C}[1]{>{\PreserveBackslash\centering}p{#1}}
\newcolumntype{R}[1]{>{\PreserveBackslash\raggedleft}p{#1}}
\newcolumntype{L}[1]{>{\PreserveBackslash\raggedright}p{#1}}
\newcommand{\specialcell}[2][c]{%
    \begin{tabular}[c]{@{}#1@{}}#2\end{tabular}}%

\title{Thinking Outside the (Gray) Box: A Context-Based Score for Assessing Value and Originality in Neural Text Generation}

\author{Giorgio Franceschelli \\
  University of Bologna \\
  \texttt{giorgio.franceschelli@unibo.it} \\\And
  Mirco Musolesi \\
  University College London \\
  University of Bologna \\
  \texttt{m.musolesi@ucl.ac.uk} \\}

\begin{document}
\maketitle
\begin{abstract}
Despite the increasing use of large language models for creative tasks, their outputs often lack diversity. Common solutions, such as sampling at higher temperatures, can compromise the quality of the results. Dealing with this trade-off is still an open challenge in designing AI systems for creativity.
Drawing on information theory, we propose a context-based score to quantitatively evaluate value and originality. This score incentivizes accuracy and adherence to the request while fostering divergence from the learned distribution. We show that our score can be used as a reward in a reinforcement learning framework to fine-tune large language models for maximum performance. We validate our strategy through experiments considering a variety of creative tasks, such as poetry generation and math problem solving, demonstrating that it enhances the value and originality of the generated solutions.
\end{abstract}

\section{Introduction}

\begin{figure}[ht]
    \centering
    \includegraphics[width=.45\textwidth]{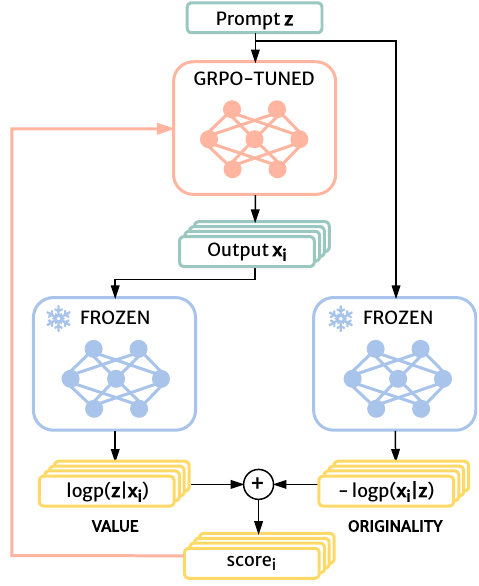}
    \caption{A summary of our method: the target model (orange) produces $G$ outputs for each prompt; a frozen reference model (blue) computes the value and originality for each output; the overall scores are used in GRPO to correct the target model (orange line).}
    \label{fig:covo}
\end{figure}

Foundation models
, particularly large language models (LLMs), are significantly transforming creative activities. 
They can serve as a foundation for co-creation systems involving human and artificial authors \cite{lin2023prompts}, can generate software code \cite{rozière2024code}, or even foster scientific research \cite{si2025can}.
However, the nature of the self-supervised learning algorithms used for the training of these models tends to make their sampling distribution as close as possible to the training data distribution \cite{franceschelli2024creativity}.
In addition, fine-tuning, such as that based on reinforcement learning from human feedback (RLHF) \cite{christiano2017deep}, is often necessary to generate appropriate and accurate responses. However, this process tends to reduce output diversity further \cite{kirk2024understanding}, and linguistic creativity tends to be lower than that of humans \cite{lu2024ai}. On the contrary, LLMs for creative tasks should produce more novel and surprising texts that maintain a high level of correctness and adherence to the request. A common solution is to sample at a higher temperature to increase diversity. However, this might lead to generating less coherent text \cite{peeperkorn2024temperature}.

To address the issues described above, we propose a new training approach for creative tasks based on \textbf{CoVO}, a \textbf{Co}ntext-based score for \textbf{V}alue and \textbf{O}riginality, to take into consideration both value and originality, i.e., two core dimensions of creativity \cite{boden2003creative,lubart1999creativity,runco2012standard}, of the generated text in the optimization of LLMs. 
The definition of CoVO is grounded in information theory \cite{mackay2003information}.
More specifically, we formulate a new optimization objective where, given a specific input, the desired output is derived by \textit{simultaneously} maximizing the conditional probability of the input given the output and minimizing the conditional probability of the output given the input under the generative model.
In this way, we optimize for solutions that are appropriate for the input request but also different from the outputs we would normally obtain from the model. In particular, we show that our information-theoretic score can be used as a reward in RL-based fine-tuning algorithms such as Group Relative Policy Optimization (GRPO) \cite{shao2024deepseekmath}, guiding pre-trained and instructed models toward more diverse yet valuable solutions (see Figure \ref{fig:covo} for a schematic overview).
Evaluations on mathematical problem solving, poetry generation, and the tasks included in \textit{NoveltyBench}~\cite{zhang2025noveltybench} demonstrate that our approach can enhance the quality and diversity of generated outputs, positioning it as a strong candidate for creativity-focused applications of current foundation models.


\section{Related Work} \label{related_work}

\subsection{Information Theory and Creativity}

The quest to provide a mathematical and computational definition of creativity has been a significant focus in recent decades. Numerous methods have been developed to define various dimensions or attributes for evaluating the creativity of AI-generated products (see, for example, \cite{franceschelli2024creativity}). 
However, these methods are often domain-specific and typically require substantial human effort to implement and assess. In contrast, solutions based on information theory \cite{shannon1948mathematical,cover1999elements} 
offer a more universally applicable approach.

Information-theoretic methods can quantify creativity by measuring the novelty and complexity of generated outputs,
without the need for extensive human intervention, making them suitable for a wide range of domains. 
Bayesian surprise \cite{baldi2010bits}, i.e., the divergence between a prior and a posterior belief, has been extensively used to measure different shades of originality, such as novelty and surprise \cite{lamb2019evaluating}. Nevertheless, \citet{varshney2019mathematical} demonstrates that there is a mathematical limit for Bayesian surprise when combined with quality measures. Surprisal \cite{tribus1961thermodynamics}, i.e., Shannon's self-information, has also been used \cite{bunescu2019learning}; \citet{barto2013novelty} extensively discuss surprisal, Bayesian surprise, and novelty. Crucially, in the context of reinforcement learning (RL), surprisal has been used as a form of intrinsic motivation to encourage the agent to explore more \cite{achiam2017surprise}. \citet{sun2025curiosity} apply this idea to improve exploration in RLHF \cite{christiano2017deep}. A similar strategy can be applied during LLM fine-tuning, either by explicitly maximizing the model’s entropy \cite{cheng2026reasoning} or by decoupling entropy and cross-entropy from KL regularization and assigning greater weight to the former \cite{slocum2025diverse}.
Additionally, mutual information has been applied to neural conversation models to improve both diversity and appropriateness \cite{li2016diversity}. 
However, all these existing approaches are not able to capture and simultaneously optimize value and originality at the same time.

\subsection{LLMs and Creativity} \label{related_work_llm}

Since the explosion of interest in LLMs in recent years, researchers have been keenly exploring their potential to exhibit creativity and ways for enhancing it \cite{franceschelli2023creativity,vadivel2026algorithmic}.
From an evaluation standpoint, research has examined whether human creativity tests can be adapted to assess the creative capabilities of LLMs \cite{stevenson2022putting}, which evaluation frameworks are most appropriate \cite{lu2026rethinking,saakyan2026death,shaib2025standardizing}, whether AI-generated texts can rival human-authored ones \cite{porter2024aigenerated,davis2024chatpgt}, and how these models may influence human creative processes \cite{doshi2024generative,lee2024empirical}.
From a generation standpoint, creativity has been addressed either at inference time or during post-training. At inference time, several decoding strategies have been proposed to improve output diversity (e.g., \cite{minh2025turning,potraghloo2025toph,troshin2025control}); prompting techniques, such as multi-perspective brainstorming \cite{lagzian2025multinovelty} or zero-shot prompting to emulate the writing styles of famous authors \cite{sawicki2023bits}, have also been explored. However, these techniques still depend on the acquired skills and are therefore constrained by the limited diversity inherent in them. Post-training approaches, instead, aim to steer the model toward richer and more diverse regions of the solution space. This can be performed through the application of RLHF \cite{christiano2017deep}, for example to write short poems that human evaluators find more creative \cite{pardinas2023leveraging}, or through RL algorithm variants \cite{anschel2025group,chung2025modifying} and additional intrinsic rewards \cite{chen2025dragrpo,dai2026cde,li2025jointly}, e.g., to promote valid but under-represented solutions \cite{he2025rewarding}, better balance exploration and exploitation \cite{cheng2026reasoning}, or optimize metrics commonly linked to creative expression \cite{ismayilzada2025creative}. However, no general method currently exists that aims to maximize diversity without also relying on external qualitative rewards.

\section{Preliminaries} \label{preliminaries}

\subsection{Language Modeling} \label{preliminaries_llms}

A $\boldsymbol{\theta}$-parameterized autoregressive language model is a probability distribution $p_{\boldsymbol{\theta}}(\mathbf{x})$ over a variable-length text sequence $\mathbf{x} = (x_1 \ldots x_T)$, where $T$ is the sequence length and each token $x_t$ is in a finite vocabulary $\mathcal{V}$ of size $N$. The probability distribution is factorized as $p_{\boldsymbol{\theta}}(\mathbf{x}) = \prod_{t=1}^T p_{\boldsymbol{\theta}}(x_t | \mathbf{x_{<t}})$, where $\mathbf{x_{<t}} = x_1 \ldots x_{t-1}$. The language model is usually trained to maximize the likelihood of the true distribution $p^*(\mathbf{x})$ for any $\mathbf{x}$ from a reference dataset (the training set). In other words, given an input $\mathbf{x_{<t}}$, the model learns to approximate the probability of each token from $\mathcal{V}$ being $x_{t}$. While this makes such a model immediately capable of scoring the probability of a given text, it also allows for the generation of new sentences. Given a conditional input (prompt) $\mathbf{z} = (z_1 \ldots z_L)$, we can decode $p_{\boldsymbol{\theta}}(\mathbf{x}|\mathbf{z})$ as the continuation of $\mathbf{z}$, i.e., through the factorized representation $p_{\boldsymbol{\theta}}(\mathbf{x} | \mathbf{z}) = \prod_{t=1}^T p_{\boldsymbol{\theta}}(x_t | \mathbf{x_{<t}}, \mathbf{z})$.

\subsection{Reinforcement Learning for Language Models} \label{preliminaries_rl}

Due to its adherence to the formal framework of Markov decision processes, RL can be used as a solution to the generative modeling problem in the case of autoregressive tasks such as text generation. 
The LLM plays the role of the agent, and each generated token represents an action $a_t$. The current version of the generated output $\mathbf{x_t}$ is part of the state $\mathbf{s_t}$ (potentially with additional information such as initial prompts). Finally, the reward $r_{t+1}$ measures the ``quality'' of the current output. A common strategy is to assign a zero reward for each $\mathbf{x}_t, t \ne T$ and a sentence-based reward when the final output is generated. Within this framework, any policy-based method can be employed to train or fine-tune the LLM to optimize a given objective.
Indeed, RL facilitates the use of non-differentiable reward functions, enabling the optimization of test-time metrics, domain-specific targets, and human preferences \cite{franceschelli2024reinforcement}. 

A widely used RL algorithm for fine-tuning LLMs is Proximal Policy Optimization \cite{schulman2017proximal}, which aims to maximize the following objective:

{\small
\begin{equation} \label{eq:ppo}
    \mathcal{J}(\boldsymbol{\theta}) = \hat{\mathbb{E}}_t \!\left[ \min(r_t(\boldsymbol{\theta}) \hat{A}_t, \mathtt{clip}(r_t(\boldsymbol{\theta}), 1 - \epsilon, 1 + \epsilon) \hat{A}_t) \right]
\end{equation}
}

\noindent where $r_t(\boldsymbol{\theta}) = \frac{\pi_{\boldsymbol{\theta}}(\mathbf{x}_t | \mathbf{z}, \mathbf{x}_{<t})}{\pi_{\boldsymbol{\theta}_{old}}(\mathbf{x}_t | \mathbf{z}, \mathbf{x}_{<t})}$ with $\pi_{\boldsymbol{\theta}}$ and $\pi_{\boldsymbol{\theta}_{old}}$ denoting the current and old policy models, repectively; $\mathbf{x}$ is the output sampled from the old policy given the prompt $\mathbf{z}$; and $\epsilon$ is a clipping factor used to stabilize training. The advantage $\hat{A}_t$ is usually computed through Generalized Advantage Estimation \cite{schulman2016highdimensional} based on the full rewards:

{\small
\begin{equation}
    R(\mathbf{z}, \mathbf{x}) = r(\mathbf{z}, \mathbf{x}) - \beta \log\frac{\pi_{\boldsymbol{\theta}}(\mathbf{x}|\mathbf{z})}{\pi_{ref}(\mathbf{x}| \mathbf{z})},
\end{equation}
}

\noindent thus integrating a KL penalty with respect to a reference model (usually, the same model before fine-tuning), and a learned value function $v_{\boldsymbol{\phi}}(\mathbf{s})$. However, learning such a value function is computationally intensive, and its training is complicated by the fact that only the last state is scored by the reward function. Moreover, the inclusion of the KL penalty as an auxiliary reward term adds complexity to the advantage estimation process. To address these issues, Group Relative Policy Optimization (GRPO) \cite{shao2024deepseekmath} has been introduced. GRPO directly adds a KL divergence term $-\beta D_{KL}(\pi_{\boldsymbol{\theta}}||\pi_{ref})$ to the loss rather than to the single rewards, and especially obviates the need for a value function approximator by using the average reward of multiple sampled outputs as the baseline:

{\small
\begin{equation}
    \hat{A}_{i,t} = \frac{r_i - \mathtt{mean}(\mathbf{r})}{\mathtt{std}(\mathbf{r})},
\end{equation}
}

\noindent with $\mathbf{r} = \{r_1, r_2, ..., r_G\}$ as the list of rewards received by each of the $G$ outputs sampled from the same prompt.

\section{A Context-Based Score for Valuable and Original Generation} \label{creativity_score}

Our goal is to derive a score that is able to quantify both value and originality at the same time.
As discussed in depth by \citet{csikszentmihalyi2014society}, creativity depends on the context in which the product is created, as the context provides the task identification and the domain information necessary to generate and validate the outcome. In turn, the output aims to solve the given task and provide a meaningful, original contribution to the current domain. Thus, our proposed score starts from mutual information, which represents a quantitative way to study the relationship between contextual, prior information, and a produced posterior outcome. More specifically, we start from the point-wise mutual information between two variables $a$ and $b$:

{\small
\begin{equation}
    I(a, b) = h(a) - h(a|b) = h(b) - h(b|a)
\end{equation}
}

\noindent with $h(y) = - \log p(y)$ and $h(x|y) = - \log(x|y)$:

{\small
\begin{equation}
    I(a, b) = \log p(a|b) - \log p(a) = \log p(b|a) - \log p(b).
\end{equation}
}

Let us now assume $a$ to be our input vector $\mathbf{z}$ and $b$ our output vector $\mathbf{x}$, obtaining:

{\small
\begin{equation}
    I(\mathbf{z},\mathbf{x}) = \log p(\mathbf{x}|\mathbf{z}) - \log p(\mathbf{x}).
\end{equation}
}

We can generalize $I(\mathbf{z},\mathbf{x})$ with two scaling factors:

{\small
\begin{equation}
    I(\mathbf{z},\mathbf{x},\lambda_1,\lambda_2) = \lambda_1 \log p(\mathbf{x}|\mathbf{z}) - \lambda_2 \log p(\mathbf{x}),
\end{equation}
}

\noindent where $I(\mathbf{z},\mathbf{x})$ is just $I(\mathbf{z},\mathbf{x},1,1)$.
Computing the \textit{absolute} probability $p(\mathbf{x})$ can be difficult, as usually generative models are developed to assign \textit{conditional} probabilities.
By applying the Bayes theorem, i.e., $\log p(a|b) = \log p(b|a) + \log p(a) - \log p(b)$, we can substitute the $\log p(\mathbf{x})$ term as follows:

{\small
\begin{equation}
    \begin{aligned}
    & I(\mathbf{z},\mathbf{x},\lambda_1,\lambda_2) = \lambda_1 \log p(\mathbf{x}|\mathbf{z}) - \lambda_2 \log p(\mathbf{x}|\mathbf{z}) \\
    &\phantom{I(\mathbf{z},\mathbf{x},\lambda_1,\lambda_2){}} - \lambda_2 \log p(\mathbf{z}) + \lambda_2 \log p(\mathbf{z}|\mathbf{x}) \\
     & = (\lambda_1 - \lambda_2) \log p(\mathbf{x}|\mathbf{z}) + \lambda_2 \log p(\mathbf{z}|\mathbf{x}) - \lambda_2 \log p(\mathbf{z}).
    \end{aligned}
\end{equation}
}

Since our goal is to find the optimal $\mathbf{x}$ for a given $\mathbf{z}$, the last term can be ignored. Moreover, we now define $\lambda_v = \lambda_2$ and $\lambda_o = \lambda_2 - \lambda_1$, thus obtaining the following objective:

{\small
\begin{equation} \label{eq:score_max_problem}
    \overline{\mathbf{x}} = \underset{\mathbf{x}}{\text{argmax}} (\lambda_v \log p(\mathbf{z}|\mathbf{x}) - \lambda_o \log p(\mathbf{x}|\mathbf{z})),
\end{equation}
}

Let us now consider the case where $\lambda_v, \lambda_o > 0$, for example, $\lambda_v = \lambda_o = 1$, which can be interpreted as the optimization of the marginal self-information $- \log p(x)$ by means of the two conditional probabilities. Solving this maximization problem involves finding the target $\mathbf{x}$ that maximizes the posterior probability of $\mathbf{z}$ while also being unlikely given $\mathbf{z}$. In other words, the optimal $\mathbf{x}^*$ must be unexpected and diverse from $p(\mathbf{x}|\mathbf{z})$, but it must also make $\mathbf{z}$ deducible from $\mathbf{x}^*$.
$- \log p(\mathbf{x}|\mathbf{z})$, commonly known as surprisal \cite{tribus1961thermodynamics}, is widely used to measure diversity and surprise \cite{barto2013novelty}, and adheres to the first requirement from the standard definition of creativity by \citet{runco2012standard}, i.e., \textit{originality} (where ``original'' is defined as ``unusual''). 
Conversely, $\log p(\mathbf{z} | \mathbf{x})$ may be interpreted as reflecting the degree to which $\mathbf{x}$ effectively addresses $\mathbf{z}$, aligning with the second requirement of the definition and a property central to creative \textit{value}.
If the request (e.g., a problem or task) can be inferred from the outcome, the latter constitutes an appropriate instance of that task or a correct, useful solution to that problem (e.g., if the request is for a sonnet or a sci-fi screenplay, the generated artifact is valuable if identified as a poem satisfying the metrical constraints of a sonnet or as a screenplay adhering to a sci-fi theme, i.e., if it maximizes the posterior probability of being recognized as a sonnet or a sci-fi screenplay, respectively).

In summary, the \textbf{CoVO} (\textbf{Co}ntext-based \textbf{V}alue and \textbf{O}riginality) score for a target $\mathbf{x}$ given a source $\mathbf{z}$ on a reference probability distribution $p$ is formally defined as:

{\small
\begin{equation} \label{eq:covo_basic_score}
    s_{CoVO}(\mathbf{z},\mathbf{x},p) = \underbrace{\lambda_v \log p(\mathbf{z}|\mathbf{x})}_\textrm{Value} \underbrace{- \lambda_o \log p(\mathbf{x}|\mathbf{z})}_\textrm{Originality}.
\end{equation}
}

\section{Implementation and Optimization with Autoregressive Models} \label{method}
We now discuss the implementation of the CoVO score with autoregressive models.
Using the notation introduced above, in the context of a $\boldsymbol{\theta}$-parameterized LLM, $p(\mathbf{x}|\mathbf{z})$ can be expressed as $\prod_{t=1}^T p_{\boldsymbol{\theta}}(x_t|\mathbf{x}_{<t},\mathbf{z})$. However, considering just the product of all the conditioned probabilities for an optimization problem would lead to preferring shorter sequences. To avoid this, we propose to use the $T$-th root: $\sqrt[T]{\prod_{t=1}^T p_{\boldsymbol{\theta}}(x_t|\mathbf{x}_{<t},\mathbf{z})}$. By leveraging the properties of the logarithm, we obtain:

{\small
\begin{equation} \label{eq:length_normalized_score}
    \lambda_v \frac{\sum_{i=1}^{|\mathbf{z}|} \log p_{\boldsymbol{\theta}}(z_i|\mathbf{z}_{<i},\mathbf{x})}{|\mathbf{z}|} - \lambda_o \frac{\sum_{j=1}^{|\mathbf{x}|} \log p_{\boldsymbol{\theta}}(x_j|\mathbf{x}_{<j},\mathbf{z})}{|\mathbf{x}|}.
\end{equation}
}

It is worth noting that the vocabulary of an LLM can be extremely large, which can cause $p_{\boldsymbol{\theta}}(a|b)$ to be small even when $a$ is the most probable event given $b$. In particular, when an LLM generates $\mathbf{x}$ given $\mathbf{z}$ and then evaluates both $p_{\boldsymbol{\theta}}(\mathbf{x}|\mathbf{z})$ and $p_{\boldsymbol{\theta}}(\mathbf{z}|\mathbf{x})$, this can lead to a significant discrepancy between the magnitude of value and originality. Since $\mathbf{x}$ has been sampled from a model based on $p_{\boldsymbol{\theta}}$, its probability would be high by definition. However, there may be various ways (possibly through synonyms) to define $\mathbf{x}$, leading to a smaller probability of $\mathbf{z}$.

Inspired by \citet{macedo2004modeling}, we propose to counteract this problem by normalizing $p_{\boldsymbol{\theta}}(a|b)$ via $n' = \frac{n - n_{min}}{n_{max} - n_{min}}$. For probabilities, $n_{min} = 0$, while $n_{max} = \max_{v \in \mathcal{V}} p_{\boldsymbol{\theta}}(v|b)$, thus obtaining the overall mapping for $p_{\boldsymbol{\theta}}$: $\frac{p_{\boldsymbol{\theta}}(x_t|\mathbf{x}_{<t},\mathbf{z})}{\max_{v \in \mathcal{V}} p_{\boldsymbol{\theta}}(v|\mathbf{x}_{<t},\mathbf{z})}$. Once again, by applying the properties of logarithms, we obtain:

{\small
\begin{equation} \label{eq:covo_advanced_score}
\begin{aligned}
    &s_{CoVO}^{AR_{norm}}(\mathbf{z},\mathbf{x},p_{\boldsymbol{\theta}}) = \lambda_v s_v(\mathbf{z}, \mathbf{x},p_{\boldsymbol{\theta}}) + \lambda_o s_o(\mathbf{z},\mathbf{x},p_{\boldsymbol{\theta}}) = \\
    &\lambda_v \frac{\sum_{i=1}^{|\mathbf{z}|} (\log p_{\boldsymbol{\theta}}(z_i|\mathbf{z}_{<i},\mathbf{x}) - \max_{v \in \mathcal{V}} \log p_{\boldsymbol{\theta}}(v|\mathbf{z}_{<i},\mathbf{x}))}{|\mathbf{z}|} -\\ 
    &\lambda_o \frac{\sum_{j=1}^{|\mathbf{x}|} (\log p_{\boldsymbol{\theta}}(x_j|\mathbf{x}_{<j},\mathbf{z}) - \max_{v \in \mathcal{V}} \log p_{\boldsymbol{\theta}}(v|\mathbf{x}_{<j},\mathbf{z}))}{|\mathbf{x}|}.
\end{aligned}
\end{equation}
}

While this normalization is rank-invariant at the token level, it subtly modifies the quantities that the score measures. In particular, the probabilities are no longer considered as absolute values, but rather relative to the highest-probability tokens, i.e., the greedy samples. In other words, the first term quantifies the degree to which $\mathbf{x}$ effectively addresses the tokens in $\mathbf{z}$ \textit{relative to how well it addresses the most likely token at each decoding step}. The second term measures how unusual the generated tokens are, on average, \textit{relative to the corresponding greedy predictions throughout the generation process}. While this normalization is not always necessary, we find it helps better balance the two score components, particularly when giving too much weight to originality would mean deviating too much from an already well-performing model (see Appendix \ref{ablation_normalization}).

Finally, calculating $p_{\boldsymbol{\theta}}(\mathbf{z}|\mathbf{x})$ is not trivial. Since LLMs are trained to complete text sequences, it is unlikely that they would generate the source text immediately after the target text (which, we should remember, is generated immediately after the source text). To address this, we consider an approximation $p_{\boldsymbol{\theta}}(\mathbf{z}|\mathbf{x}')$, where $\mathbf{x}' = \mathbf{x} + \mathbf{q}$. Here, $\mathbf{q}$ represents an additional question, such as ``How would you describe this text?'' or a similar formulation designed solely to increase the likelihood of generating the source text $\mathbf{z}$ (as well as alternative sources). Appendix \ref{ablation_prompt} reports an ablation study on the low sensitivity of different $\mathbf{q}$ formulations for the computation of the final score.

Once the CoVO score has been defined, its adoption in an RL framework is straightforward. As previously introduced, we can directly utilize our CoVO score as the final reward for the generated sequence. 
Then, the model can be trained with any policy gradient method. Our experiments leverage GRPO \cite{shao2024deepseekmath}, which is a state-of-the-art choice for training language models, with and without the per-token KL divergence loss.

\section{Experiments} \label{experiments}

We evaluate the effectiveness of our RL strategy through three case studies: poetry generation, mathematical problem resolution, and the tasks included in NoveltyBench\footnote{The code and results of the experiments can be found at: \\ \url{https://anonymous.4open.science/r/CoVO-grpo/}}.
In all experiments, we employ two settings, i.e., GRPO to maximize the score from Equation \ref{eq:covo_advanced_score} \textit{without} the KL divergence loss, i.e., with $\beta = 0.0$ (CoVO); and GRPO to maximize the score from Equation \ref{eq:covo_advanced_score} \textit{with} the KL divergence loss, i.e., with $\beta = 0.05$ (CoVO+KL). Both methods assume $\lambda_v = \lambda_o = 1.0$. Finally, we restrict our baselines to the original model and, if available, a model tuned solely on the environmental reward. While other methods have been proposed recently to enhance the diversity of generated outputs (as introduced in Section \ref{related_work_llm}), to the best of our knowledge, none of them can be directly applied without an extrinsic reward, thus being incompatible with the majority of our experiments where no environmental reward is available. In addition, while it is common to induce diversity at the sampling level \cite{peeperkorn2024temperature}, our approach is orthogonal to the chosen sampling strategy and can benefit from more advanced methods as well. On the contrary, our evaluation setting focuses on verifying whether our reward scheme can increase value and originality, regardless of how the output is sampled.

\subsection{Poetry Generation} \label{exp_poetry_generation}

\begin{table*}[ht]
\centering
\resizebox{\textwidth}{!}{%
\begin{tabular}{|l|cccc|cccc|}
 \hline
 & \multicolumn{4}{c|}{In-distribution} & \multicolumn{4}{c|}{Out-of-distribution} \\
 \cline{2-9}
 \multirow{-2}{*}{Method} & Corr. $\uparrow$ & Metric (L/S) $\uparrow$ & T-LCS $\downarrow$ & Tone $\uparrow$ & Corr. $\uparrow$ & Metric (L/S) $\uparrow$ & T-LCS $\downarrow$ & Tone $\uparrow$  \\
 \hline
 \texttt{Meta-Llama-3-8B} & $0.944$ & $0.360$ / $0.162$ & $9.640$ / $149$ & $0.634_{\pm 0.055}$ & $0.960$ & $0.413$ / $0.099$ & $6.288$ / $68$ & $0.610_{\pm 0.054}$ \\
 + CoVO                   & $0.928$ & $0.080$ / $0.061$ & $4.712$ / \textcolor{white}{00}$8$ & $0.662_{\pm 0.051}$ & $0.904$ & $0.240$ / $0.056$ & $4.704$ / $10$ & $0.637_{\pm 0.044}$ \\
 + CoVO + KL              & $0.936$ & $0.300$ / $0.132$ & $5.072$ / \textcolor{white}{0}$16$ & $0.661_{\pm 0.049}$ & $0.992$ & $0.347$ / $0.113$ & $4.992$ / \textcolor{white}{0}$9$ & $0.667_{\pm 0.047}$ \\
 \hline
 \texttt{Qwen-2.5-3B}     & $0.992$ & $0.520$ / $0.102$ & $5.112$ / \textcolor{white}{00}$8$ & $0.817_{\pm 0.041}$ & $0.976$ & $0.600$ / $0.096$ & $4.856$ / \textcolor{white}{0}$8$ & $0.732_{\pm 0.049}$ \\
 + CoVO                   & $0.960$ & $0.440$ / $0.126$ & $5.032$ / \textcolor{white}{00}$7$ & $0.744_{\pm 0.051}$ & $0.984$ & $0.493$ / $0.092$ & $4.888$ / \textcolor{white}{0}$7$ & $0.694_{\pm 0.050}$ \\
 + CoVO + KL              & $0.984$ & $0.440$ / $0.115$ & $5.056$ / \textcolor{white}{00}$7$ & $0.779_{\pm 0.048}$ & $0.984$ & $0.507$ / $0.093$ & $4.840$ / \textcolor{white}{0}$8$ & $0.690_{\pm 0.053}$ \\
 \hline
\end{tabular}
}
\caption{Aggregate results over 5 seeds of generated poems from training prompts (left) and testing prompts (right). Scores on poetical metrics are reported at the line (L) and syllable (S) levels and only consider requests for styles with specific metrical properties. Under T-LCS, we report both the mean and the maximum longest common substring across all generated poems. The mean and the 95\% confidence interval are reported for tone adherence.\label{poetry_generation_avg_score}}
\end{table*}

\begin{figure}[ht]
    \centering
    \includegraphics[width=0.48\textwidth]{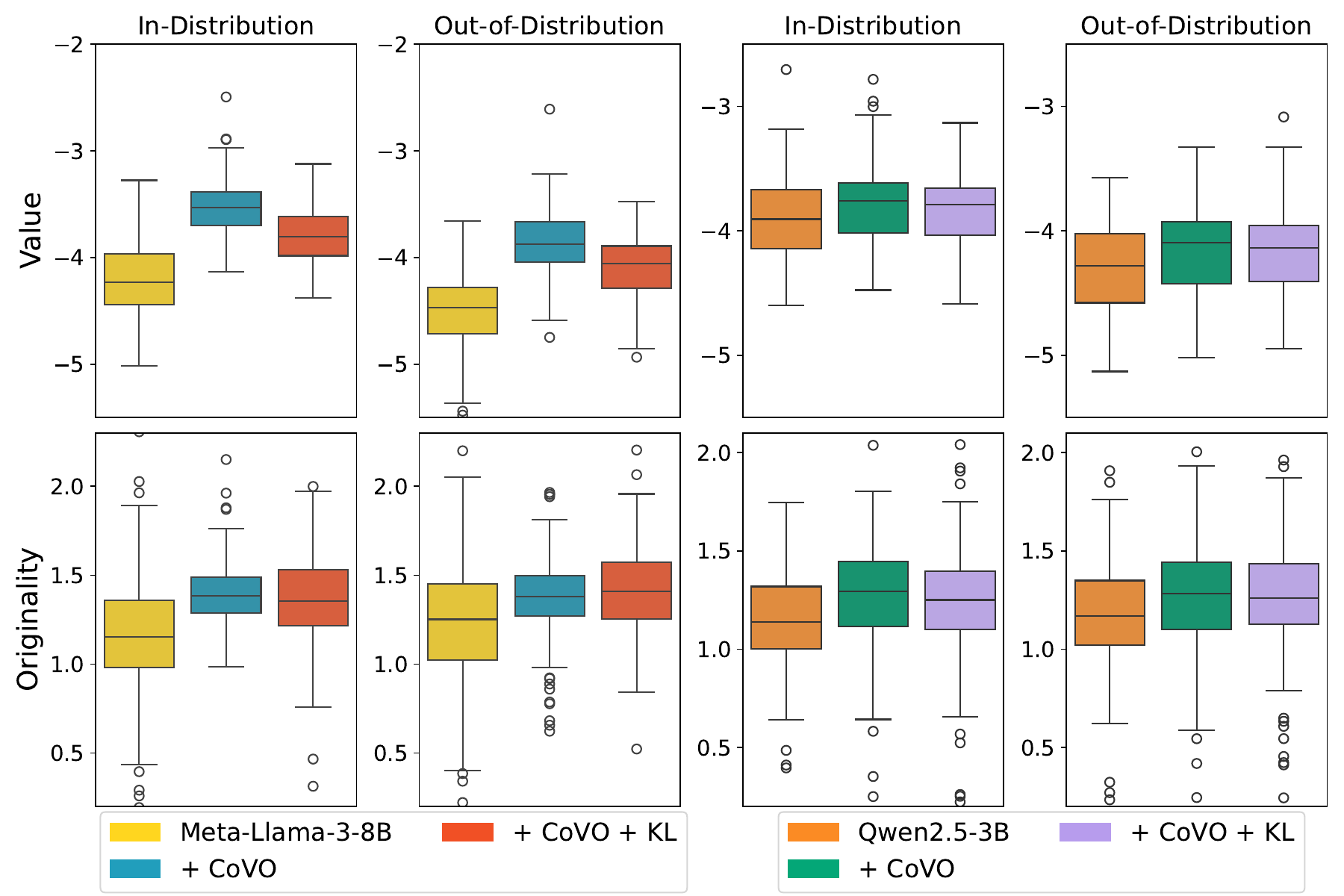}
    \caption{The distribution of value and originality scores for the in-distribution and out-of-distribution poems generated by the original models and our tuned versions.}
    \label{fig:val_orig}
\end{figure}

\textbf{Experimental Setup.}
The first set of experiments concerns a very common creative task, aiming to teach the LLM to generate poems that are both original and valuable.
More specifically, we follow the approach outlined by \citet{bradley2024quality} and instruct the model to write a poem in a particular style and tone. We adopt the \texttt{Meta-Llama-3-8B} \cite{llamateam2024llama} and \texttt{Qwen2.5-3B} \cite{qwen2025qwen25} models as our pre-trained agents. Since we do not use the instruction-tuned models, we prompt them with a few-shot examples of the task to make them more likely to produce the desired output in the desired form (the full prompt is reported in Appendix \ref{implementation_details}, together with the training parameters). Instead of fine-tuning the entire network, we use Low-Rank Adaptation (LoRA) \cite{hu2022lora}.
The original model is also used to compute the score.
All sampling happens with top-$k$ ($k = 50$) and at a temperature of $1.0$.
We perform a quantitative evaluation where we compute poetical metrics for quality (lexical correctness of poems, adherence to line- and syllable-level constraints, and tone adherence to the request through zero-shot classification with \texttt{bart-large-mnli} model \cite{lewis2020bart}) and for originality (accidental reproduction of existing poems). For the latter, we define a Token-based Longest Common Substring (T-LCS) score and use it to compare generated poems with a reference dataset of approx. 84k public-domain poems extracted from Project Gutenberg (please refer to Appendix \ref{gutenverse_dataset} for a first presentation of our GutenVerse dataset). While a generated poem can be an accidental reproduction of a protected work or a different source (e.g., a song), we believe it can provide a useful evaluation tool to understand the general degree of originality.

\noindent\textbf{Experimental Results.}
Table \ref{poetry_generation_avg_score} reports the scores about the compliance of poetical constraints at the syllable and line levels, lexical correctness (the ratio of poems not containing noisy text), tone adherence (the zero-shot classification of that poem being of that tone rather than its opposite), and accidental reproduction rate (the mean and maximum token-based longest common substring).
Overall, our CoVO-based fine-tuning leads to higher tone adherence for Llama-based models and a lower reproduction rate for both models, at the potential cost of metric adherence, especially without the KL loss. Indeed, its role seems to foster quality (especially in terms of metrical correctness), trading off some originality. On the contrary, not using the KL loss arguably avoids any significant reproduction, as demonstrated by the very low maximum token-based longest common substring.
Interestingly, these considerations align well with our CoVO score. Figure \ref{fig:val_orig} reports the value and originality according to Equation \ref{eq:covo_advanced_score} under the pre-trained model.
While the two methods do not differ significantly from the baseline (possibly due to opposing forces of value and originality \cite{varshney2019mathematical}), we again see that the presence of KL leads to slightly lower scores, even though its effect varies slightly across the two tested models.
However, aggregated scores, such as those presented here, might be insufficient. For a more complete overview, we also conducted a fine-grained analysis of the generated poems in Appendix \ref{poem_analysis}.

\begin{table*}[ht]
\centering
\resizebox{\textwidth}{!}{%
\begin{tabular}{|L{3.2cm}||C{2.1cm}|C{2.cm}|C{1.4cm}C{1.4cm}||C{2.2cm}|C{2.cm}|C{1.4cm}C{1.4cm}|} 
\hline
& \multicolumn{4}{c||}{GSM8K} & \multicolumn{4}{c|}{MATH} \\
\cline{2-9}
& & & \multicolumn{2}{c||}{Cross-Input Diversity} & & & \multicolumn{2}{c|}{Cross-Input Diversity} \\
\multirow{-3}{*}{\parbox{3cm}{\texttt{MetaMath-\\Mistral-7B}}} & \multirow{-2}{*}{Accuracy} & \multirow{-2}{*}{Entropy} & EAD & SBERT & \multirow{-2}{*}{Accuracy} & \multirow{-2}{*}{Entropy} & EAD & SBERT \\
\hline
Original model    & $77.96\% \;(3)$ & $0.156_{\pm .002}$ & $2.0071$ & $0.6402$ & $33.55\% \;(483)$ & $0.185_{\pm .002}$ & $5.7239$ & $\mathbf{0.8032}$ \\
+ Ext             & $78.18\% \;(4)$ & $0.154_{\pm .002}$ & $2.0045$ & $\mathbf{0.6404}$ & $33.19\% \;(469)$ & $0.182_{\pm .002}$ & $5.7187$ & $0.8027$ \\
+ Ext + KL        & $78.08\% \;(5)$ & $0.154_{\pm .002}$ & $2.0077$ & $0.6401$ & $33.56\% \;(476)$ & $0.182_{\pm .002}$ & $5.7517$ & $0.8028$ \\
+ CoVO            & $78.12\% \;(3)$ & $\mathbf{0.164_{\pm .003}}$ & $\mathbf{2.0509}$ & $0.6403$ & $33.29\% \;(521)$ & $\mathbf{0.197_{\pm .002}}$ & $5.8219$ & $0.8029$ \\
+ CoVO + KL       & $78.12\% \;(3)$ & $\mathbf{0.163_{\pm .003}}$ & $2.0464$ & $0.6402$ & $32.79\% \;(533)$ & $\mathbf{0.195_{\pm .002}}$ & $\mathbf{5.8442}$ & $0.8030$ \\
+ CoVO + Ext      & $\mathbf{78.33}\% (4)$ & $0.158_{\pm .002}$ & $2.0340$ & $\mathbf{0.6404}$ & $\mathbf{33.76}\% (503)$ & $0.188_{\pm .002}$ & $5.8114$ & $0.8030$ \\
+ CoVO + Ext + KL & $77.95\% \;(4)$ & $0.157_{\pm .002}$ & $2.0367$ & $0.6402$ & $33.74\% \;(497)$ & $0.187_{\pm .002}$ & $5.8082$ & $0.8031$ \\
\hline
\end{tabular}
}
\caption{Accuracy and diversity for the GSM8K (left) and MATH (right) test sets. In brackets, the number of responses exceeding the maximum token limit. The mean and the 95\% confidence interval are reported for sequence-level entropy.
\label{math_covo_rl_mistral}}
\end{table*}

\subsection{Math Problem Resolution} \label{exp_math_resolution}

\textbf{Experimental Setup.}
The second set of experiments concerns a more practical and quantitative task, as it aims to teach the LLM to solve mathematical problems through more diverse procedures. In particular, we focus on the Mistral-based \texttt{MetaMath-Mistral-7B} model, i.e., fine-tuned with self-supervised learning on the MetaMathQA \cite{yu2024metamath}. It is a dataset of textual math questions paired with responses where the numerical answer is easily separable from the textual procedure. While the entire set contains 395k entries, making an additional training epoch too expensive, MetaMathQA is composed of entries from two different training sets, then augmented with various techniques: GSM8K \cite{cobbe2021training} and MATH \cite{hendrycks2021measuring}. Since we are only interested in the questions, we limit our training to a single epoch over those datasets. Moreover, we exclude all questions with a tokenized length of either question or answer greater than 512, obtaining 14876 out of 14973 total entries.
We separate the procedure and the answer from each solution to train our model and use the numerical answer to check the correctness of the predicted solution.
The RL problem can then be formulated considering up to two rewards: our CoVO score computed on the procedure and a verifiable, extrinsic reward based on the correctness of the answer. Instead of fine-tuning the entire model, we adopt a more parameter-efficient strategy with LoRA and use the original model to perform the CoVO score computation. 
Following \citet{yu2024metamath}, the outputs are obtained with a greedy strategy.
The evaluation considers both GSM8K and MATH test sets (limited to the entries with a tokenized length of question and answer smaller than 512, i.e., all 1319 entries for GSM8K and 4546 out of 5000 for MATH). 
We compute the percentage of correct solutions together with three diversity metrics: expectation-adjusted distinct N-grams (EAD) \cite{liu2022rethinking} and sentence embedding cosine similarity (SBERT) \cite{reimers2019sentence}, which should measure syntactical and semantical diversity, respectively \cite{kirk2024understanding}, plus entropy. EAD counts the number of distinct N-grams (averaging over $N=1 \ldots 5$) across all generated responses and removes the bias toward shorter inputs by scaling the number of distinct tokens based on their expectations. The SBERT metric computes the average of the cosine similarity between the embeddings of any possible pairs of outputs and returns 1 minus the similarity. 
While this was originally based on Sentence-BERT, we employ the more recent \texttt{all-mpnet-base-v2} instead, as suggested by their developers\footnote{See the developers’ note about the original Sentence-BERT model and recommendations for alternative models at: \url{https://huggingface.co/sentence-transformers/bert-large-nli-stsb-mean-tokens}}.
Following \citet{kirk2024understanding}, we compute \textit{cross-input} EAD and SBERT, i.e., we derive them by considering all outputs produced for a specific seed together. 
Finally, we compute the sequence-level entropy of the generated outputs to evaluate the potential diversity of the learned probability distributions.

\noindent \textbf{Experimental Results.}
Table \ref{math_covo_rl_mistral} reports the results for the GSM8K and MATH test sets. For the GSM8K test set, while all methods achieve similar results, using the CoVO score only (with and without the KL loss) leads to higher entropy and greater EAD diversity. In contrast, the presence of the math reward leads to higher accuracy, especially when KL is absent.
The results for the MATH test set confirm that using the CoVO score alone increases diversity, and that the most accurate method is the one trained to optimize the CoVO score and the extrinsic reward, with a negligible trade-off in terms of diversity, since both the entropy and the cross-input EAD are still substantially higher than those of the baselines. However, removing the extrinsic reward likely pushes the model too far from its pre-trained version, causing the accuracy to decrease.


\begin{table}[t]
\centering
\resizebox{.48\textwidth}{!}{%
\begin{tabular}{|l|cc|}
 \hline
 Method & Quality & Novelty \\
 \hline
 \texttt{Llama3.2-3B-Instruct} & $3.90 \pm 0.16$ & $4.31 \pm 0.23$ \\
 + CoVO                        & $3.89 \pm 0.20$ & $\mathbf{8.71 \pm 0.16}$ \\
 + CoVO + KL                   & $\mathbf{4.88 \pm 0.19}$ & $7.13 \pm 0.23$ \\
 \hline
 \texttt{Qwen-2.5-7B-Instruct} & $3.51 \pm 0.16$ & $3.62 \pm 0.23$ \\
 + CoVO                        & $\mathbf{3.91 \pm 0.18}$ & $\mathbf{4.52 \pm 0.24}$ \\
 + CoVO + KL                   & $\mathbf{3.85 \pm 0.17}$ & $\mathbf{4.31 \pm 0.23}$ \\
 \hline
\end{tabular}
}
\caption{The utility (quality) and distinct (novelty) scores on NoveltyBench (curated partition) for the original models and our methods (tuned on the wildchat partition). Mean and 95\% confidence intervals are reported over 5 seeds.
\label{noveltybench_table}}
\end{table}

\subsection{NoveltyBench}

\textbf{Experimental Setup.}
Finally, we also experiment with NoveltyBench \cite{zhang2025noveltybench}, a recent benchmark that aims to evaluate the ability of language models to produce multiple distinct and high-quality outputs. NoveltyBench contains two sets of prompts thought for eliciting diverse responses: the `curated' partition, with $100$ prompts manually curated by the paper's authors, and the `wildchat' partition, with $1000$ prompts sourced from the WildChat-1M dataset \cite{zhao2024wildchat}. The benchmark requires a language model to generate $10$ outputs per prompt using a temperature setting of $1.0$, after which it computes scores for novelty and quality. 
To assess novelty, the $10$ outputs are grouped into equivalence classes using a fine-tuned DeBERTa model \cite{he2021deberta}, and the number of distinct classes is reported as the novelty score. To assess quality, it computes the cumulative utility of the 10 outputs, where the utility is $0$ if the $i$-th output has the same equivalence class as a preceding output, and a calibrated reward from \texttt{Skywork-Reward-Gemma-2-27B-v0.2} \cite{liu2024skywork} otherwise (we refer to \citet{zhang2025noveltybench} for full details).
We fine-tune the \texttt{Qwen2.5-7B-Instruct} \cite{qwen2025qwen25} and the \texttt{Llama-3.2-3B-Instruct} \cite{llamateam2024llama} models for a single epoch on the `wildchat' partition (the full parameters are reported in Appendix \ref{implementation_details}); then, we compute the novelty and quality scores on the `curated' partition, and we compare them with the performance of the original model.

\noindent \textbf{Experimental Results.}
Table \ref{noveltybench_table} reports the quality and novelty scores achieved by our methods, compared to those from the original instructed models. For \texttt{Llama-3.2-3B-Instruct}, which already obtains good results, optimizing for the CoVO score alone results in substantial improvement in novelty, without any significant variations in quality. Adding the KL penalty, which keeps the model closer to the original version, trades off some of the gained novelty (while still improving it substantially) for a substantial improvement in quality. On the other hand, \texttt{Qwen-2.5-7B-Instruct} represents a lower starting point. Because of this, optimizing for the CoVO score improves both quality and novelty metrics, while adding the KL penalty reduces these gains by keeping the model closer to its original version (while still significantly improving upon it).

\section{Conclusion} \label{conclusion}
In this paper, we presented CoVO, a novel score that quantifies the value and originality of neurally-generated text. The definition of CoVO is theoretically grounded on information theory and includes the conditional probability of the model’s outputs given the inputs, and vice versa. We also proposed an optimization problem where a generative model aims to maximize this score to generate more creative products, and detailed how to use it in language modeling. We conducted experiments on poetry generation, math problem solving, and tasks from NoveltyBench, exploring trade-offs in accuracy vs diversity.

Effectively balancing value and originality maximization remains an open question, but our experimental results appear to show that our score is able to capture properties functionally aligned with creativity-relevant aspects of language generation, as demonstrated through both conceptual analysis and empirical results. When used as a reward function in GRPO, it promotes improvements in both the novelty and quality of model outputs, even with relatively few optimization steps, and without necessarily relying on external rewards, which are required by most alternative methods.

Our research agenda aims to extend our method to other models and tasks, to include inference-level strategies such as creativity-oriented sampling schemes, and to explore its use for evaluation (e.g., in a Best-of-N setting \cite{stiennon2020learning}) rather than solely for optimization. We also plan to investigate the definition of additional scores for capturing other potentially relevant aspects of the creative process. Despite being costly and inherently constrained \cite{davis2024chatpgt}, assessing whether our creativity score aligns with human judgment is another key direction for future work.

\section*{Limitations}

There are a few limitations worth noting. Firstly, our score represents only a quantifiable approximation of a particular theoretical perspective on creativity, grounded in the dimensions of value and originality. For example, value has been considered solely from the perspective of effectiveness, while other dimensions have been proposed as well (e.g., interestingness \cite{boden1994dimensions} or monetary worth \cite{lepak2007value}). Quality and correctness are two additional key dimensions of creative value \cite{boden2003creative}. Although they can be indirectly incorporated through extrinsic environmental rewards (as in math problem-solving tasks) and the KL-penalty loss that preserves pre-training behaviors, they are not explicitly captured by our score. In addition, other classic definitions of creativity, such as that presented by \citet{boden2003creative}, add a third requirement by splitting originality into novelty and surprise. Moreover, our score reflects a specific view of the evaluation of creativity based on the generated outputs and does not account for potential alternative theories (for example, arising from different cultures \cite{lubart1999creativity}) and perspectives \cite{rhodes1961analysis}. From a practical point of view, computing the CoVO score requires an additional pass inside the model, which represents a potential bottleneck for larger LLMs: for the math problem-solving case study, introducing the CoVO-based reward in addition to the external reward leads to an overhead of ~15 seconds per iteration, which is equal to a 50\% overhead on the overall training time. In addition, maximizing it can lead to reward hacking \cite{pan2022the}, where one of the two score components is exploited, e.g., only maximizing originality (outputting something completely unpredictable and meaningless) or only maximizing value (e.g., repeating the request, which is not original at all and helps hack the value-based reward function). We experienced a few of these failure modes when the CoVO score was the only optimization target (please refer to the detailed analysis in Appendix \ref{poem_analysis}). However, retaining both scores and incorporating the KL divergence largely mitigated these failure modes in our experiments.
Finally, our evaluation is currently limited to only three relatively short-form text generation tasks. While their generalizability is supported by the theoretical framework discussed above, the resulting performance was experimentally evaluated for a finite number of scenarios.

\section*{Ethical Considerations}

The authors are aware of the potential impact that generative technologies might have on the production of artistic outputs and, as a consequence, on human artists. This work may contribute to enhancing the quality of generated outputs. However, the authors argue that typical traits of human creativity, such as the active participation of artists in the creative process, cannot be directly replicated by machines, as pointed out by Runco in the updated standard definition of creativity \cite{runco2025updating}. The authors refer interested readers to a previous work of theirs (\textit{citation removed for double-blind submission}), in which these themes are discussed in detail.

\bibliography{biblio}

\clearpage

\appendix

\section{Implementation Details} \label{implementation_details}

\noindent \textbf{Computational and Experimental Setup.} The experiments were carried out using a Linux-based local server with two 80GB NVIDIA H100 GPUs, running Python 3.11.9. All the training was conducted with a random seed equal to 1 (set through the \texttt{set\_seed} method from the HuggingFace \texttt{transformers} library), while poems and NoveltyBench outputs were sampled at inference time with five different seeds (0, 1, 42, 121, and 12345). The hyperparameters were chosen to maximize the efficient use of the available computational resources, while all remaining parameters were kept at their default values from the HuggingFace \texttt{transformers} and \texttt{trl} libraries. Only the learning rates were tuned for the different tasks according to training performance.

\noindent \textbf{Poetry Generation.} Table \ref{tab:covo_rl_poetry_params} reports the full training parameters for the experiments on poetry generation. 
\begin{table}[th]
    \centering    

    \begin{tabular}{lc}
         \hline \noalign{\vskip 1mm}
        \textbf{Parameter} & \textbf{Value} \\ [0.5ex]
        \hline
        Total batches & 100 \\
        Batch size $B$ & 4 \\
        Gradient accumulation steps & 8 \\
        Max new tokens & 256 \\
        Temperature & 1. \\
        Top-$k$ & 0 \\
        Optimizer & Adam \\
        Learning rate & 1e-5 \\
        Max gradient normalization & 100. \\
        Rank (LoRA) & 16 \\
        $\alpha$ parameter (LoRA) & 32 \\
        Dropout (LoRA) & 0.05 \\
        Training iterations & 1 \\
        Scale rewards & True \\
        $\beta$ (when used) & 0.05 \\
        Number of generations $G$ & 4 \\
    \end{tabular}
    \caption{Training parameters for poetry generation used in the experiments.}
    \label{tab:covo_rl_poetry_params}
\end{table}

\noindent The prompt for generation leverages \textit{Nothing gold can stay} by Robert Frost, \textit{Fame is a bee} by Emily Dickinson, and \textit{Epitaph} by William Carlos Williams for few-shot learning:

{\small
\begin{tcolorbox}[colback=white,colframe=black]
Write a fatalistic epigram poem of high, award winning quality.\\
\\
Nature’s first green is gold,\\
Her hardest hue to hold.\\
Her early leaf’s a flower;\\
But only so an hour.\\
Then leaf subsides to leaf.\\
So Eden sank to grief,\\
So dawn goes down to day.\\
Nothing gold can stay.\\
\\
\\
Write an ironic quatrain poem of high, award winning quality.\\
\\
Fame is a bee.\\
It has a song-\\
It has a sting-\\
Ah, too, it has a wing.\\
\\
\\
Write a naturalistic epitaph poem of high, award winning quality.\\
\\
An old willow with hollow branches\\
Slowly swayed his few high fright tendrils\\
And sang:\\
\\
Love is a young green willow\\
Shimmering at the bare wood's edge.\\
\\  
\\
Write a \{\texttt{tone}\} \{\texttt{style}\} poem of high, award winning quality.
\end{tcolorbox}
}

\noindent In addition, the training phase includes requests with tone-style pairs sampled among `dark', `happy', `mysterious', `reflective' or `romantic' for the tone, and `ballad', `haiku', `hymn', `limerick' or `sonnet' for the style, as done by \citet{bradley2024quality}. At inference time, we also consider `cinquain', `couplet', `free verse', `ode' or `tanka' as styles and `cutting', `nostalgic', `poignant', `solemn' or `whimsical' as tones. Instead, the prompt used for computing $p_{\boldsymbol{\theta}}(\mathbf{z}|\mathbf{x})$ is:

{\small
\begin{tcolorbox}[colback=white,colframe=black]
Try to guess what instruction led to the following poem:\\
\\
\#\#\# Poem:\\
\{\texttt{poem}\}\\
\\
\#\#\# Instruction: Write a
\end{tcolorbox}
}

\noindent with ``\{\texttt{tone}\} \{\texttt{style}\} poem of high, award winning quality'' as the target completion $\mathbf{z}$. Finally, the zero-shot classification for the tone adherence is performed with the given tone and `not' plus the given tone as the candidate labels (e.g., if the required tone is `happy', the two labels are [`happy', `not happy']).

\begin{table}[ht]
    \centering    
    \begin{tabular}{lc}
        \hline \noalign{\vskip 1mm}
        \textbf{Parameter} & \textbf{Value} \\ [0.5ex]
        \hline
        Total epochs & 1 \\
        Batch size $B$ & 4 \\
        Gradient accumulation steps & 8 \\
        Max new tokens & 512 \\
        Temperature & 1. \\
        Top-$k$ & 0 \\
        Optimizer & Adam \\
        Learning rate & 1e-6 \\
        Max gradient normalization & 100. \\
        Rank (LoRA) & 16 \\
        $\alpha$ parameter (LoRA) & 32 \\
        Dropout (LoRA) & 0.05 \\
        Training iterations & 1 \\
        Scale rewards & True \\
        $\beta$ (when used) & 0.05 \\
        Number of generations $G$ & 4 \\
        Reward for correct answer & +1. \\
    \end{tabular}
    \caption{Training parameters for math problem solving.}
    \label{tab:covo_rl_math_params}
\end{table}

\begin{table}[ht]
    \centering    
    \begin{tabular}{lc}
        \hline \noalign{\vskip 1mm}
        \textbf{Parameter} & \textbf{Value} \\ [0.5ex]
        \hline
        Total epochs & 1 \\
        Batch size $B$ & 4 \\
        Gradient accumulation steps & 8 \\
        Max new tokens & 512 \\
        Temperature & 1. \\
        Top-$k$ & 0 \\
        Optimizer & Adam \\
        Learning rate & 1e-4 \\
        Max gradient normalization & 100. \\
        Rank (LoRA) & 16 \\
        $\alpha$ parameter (LoRA) & 32 \\
        Dropout (LoRA) & 0.05 \\
        Training iterations & 1 \\
        Scale rewards & True \\
        $\beta$ (when used) & 0.05 \\
        Number of generations $G$ & 4 \\
    \end{tabular}
    \caption{Training parameters for NoveltyBench.}
    \label{tab:covo_rl_nb_params}
\end{table}

\noindent \textbf{Math Problem Solving.} Table \ref{tab:covo_rl_math_params} reports the full training parameters for math problem resolution. We also adopted the same two different prompts from \citet{yu2024metamath}, i.e.:

{\small
\begin{tcolorbox}[colback=white,colframe=black]
Below is an instruction that describes a task. Write a response that appropriately completes the request.\\
\\
\#\#\# Instruction:\\
\{\texttt{question}\}\\
\\
\#\#\# Response:
\end{tcolorbox}
}

\noindent at training time, and:

{\small
\begin{tcolorbox}[colback=white,colframe=black]
Below is an instruction that describes a task. Write a response that appropriately completes the request.\\
\\
\#\#\# Instruction:\\
\{\texttt{question}\}\\
\\
\#\#\# Response: Let's think step by step.
\end{tcolorbox}
}

\noindent at inference time. Instead, in oder to derive the value of $p_{\boldsymbol{\theta}}(\mathbf{z}|\mathbf{x})$ we used the prompt reported below:

{\small
\begin{tcolorbox}[colback=white,colframe=black]
Below is a response that appropriately completes a request. Write the instruction that describes the task.\\
\\
\#\#\# Response:\\
\{\texttt{response}\}\\
\\
\#\#\# Instruction:
\end{tcolorbox}
}

\noindent \textbf{NoveltyBench.} Table \ref{tab:covo_rl_nb_params} reports the full training parameters for NoveltyBench. 
At training and inference time, we simply adopt the following prompt:

{\small
\begin{tcolorbox}[colback=white,colframe=black]
user\\
\\
\{\texttt{prompt}\}\\
assistant
\end{tcolorbox}
}

\noindent where \texttt{user} and \texttt{assistant} are keywords used by the model to identify different roles in the chat. Instead, for computing $p_{\boldsymbol{\theta}}(\mathbf{z}|\mathbf{x})$ we used the following:

{\small
\begin{tcolorbox}[colback=white,colframe=black]
Below is a response that appropriately solves a task. Write the instruction that describes the task.\\
\\
\#\#\# Response:\\
\{\texttt{response}\}\\
\\
\#\#\# Instruction:
\end{tcolorbox}
}

\section{Ablation Study on the Max-Probability Normalization} \label{ablation_normalization}

The final formulation of the CoVO score adopts a max-probability normalization for both the value and originality scores, as defined in Equation \ref{eq:covo_advanced_score}. However, as discussed in Section \ref{method}, this transformation slightly changes what the score measures, namely the relative effectiveness of the output in addressing the input and the relative originality of the output given the input, rather than their absolute counterparts. While the primary motivation is numerical stability, we also study the practical effect of this transformation.

\begin{figure*}[ht]
    \centering
    \includegraphics[width=0.98\textwidth]{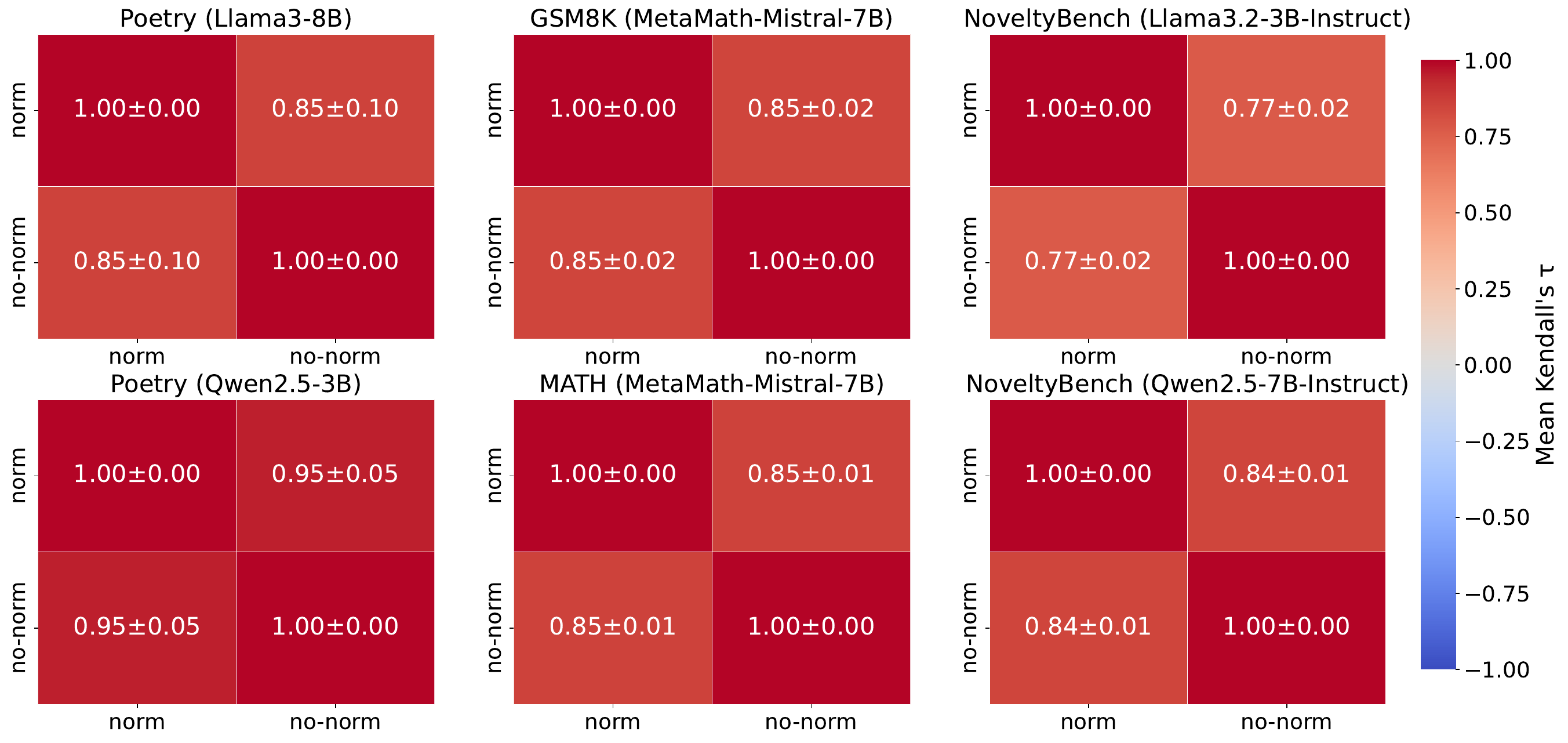}
    \caption{Correlation matrices between the CoVO scores for our three case studies with and without the max-normalization step. Kendall's $\tau$ mean and 95\% confidence interval between scores for multiple solutions over the same inputs are reported.}
    \label{fig:correlation_norm}
\end{figure*}

\begin{table}[ht]
\centering
\resizebox{.48\textwidth}{!}{%
\begin{tabular}{|L{3.8cm}|C{2.0cm}|C{2.0cm}|}
 \hline
 Experiment & Value & Originality \\
 \hline
 \multicolumn{3}{|c|}{Poetry generation  - \texttt{Llama3-8B}} \\
 \hline
 with normalization    & $-4.30_{\pm 0.05}$ & $0.80_{\pm 0.16}$ \\
 without normalization & $-5.35_{\pm 0.07}$ & $1.65_{\pm 0.32}$ \\
 \hline
 \multicolumn{3}{|c|}{Poetry generation  - \texttt{Qwen-2.5-3B}} \\
 \hline
 with normalization    & $-3.82_{\pm 0.06}$ & $3.15_{\pm 0.22}$ \\
 without normalization & $-4.72_{\pm 0.07}$ & $4.79_{\pm 0.27}$ \\
 \hline
 \multicolumn{3}{|c|}{Math problem solving  - GSM8K} \\
 \hline
 with normalization    & $-1.17_{\pm 0.01}$ & $0.72_{\pm 0.01}$ \\
 without normalization & $-1.32_{\pm 0.01}$ & $1.02_{\pm 0.01}$ \\
 \hline
 \multicolumn{3}{|c|}{Math problem solving  - MATH} \\
 \hline
 with normalization    & $-0.91_{\pm 0.01}$ & $0.50_{\pm 0.01}$ \\
 without normalization & $-1.06_{\pm 0.02}$ & $0.74_{\pm 0.01}$ \\
 \hline
 \multicolumn{3}{|c|}{NoveltyBench  - \texttt{Llama3.2-3B-Instruct}} \\
 \hline
 with normalization    & $-1.88_{\pm 0.02}$ & $0.37_{\pm 0.01}$ \\
 without normalization & $-2.41_{\pm 0.03}$ & $0.79_{\pm 0.02}$ \\
 \hline
 \multicolumn{3}{|c|}{NoveltyBench  - \texttt{Qwen-2.5-7B-Instruct}} \\
 \hline
 with normalization    & $-2.87_{\pm 0.04}$ & $0.52_{\pm 0.01}$ \\
 without normalization & $-3.19_{\pm 0.04}$ & $0.98_{\pm 0.02}$ \\
 \hline
\end{tabular}
}
\caption{The value and originality scores with and without the max-probability normalization step. Mean $\pm$ 95\% confidence intervals are reported over 4 generated outputs from the entire training set for poetry and NoveltyBench and from a sample of 1000 training data for GSM8K and MATH.
\label{covo_scores_unnorm}}
\end{table}

First, we compare the impact of keeping or removing the normalization when scoring different outputs for the same input. More specifically, we compute the mean and 95\% confidence interval of Kendall's $\tau$ correlation among $G=4$ generated outputs over different inputs to evaluate whether normalization changes the relative ranking of alternative solutions (as generated during GRPO tuning). We consider all 25 training pairs for poetry, a sample of 1,000 training questions from the GSM8K and MATH datasets, and the entire \texttt{wildchat} partition for NoveltyBench. As shown in Figure \ref{fig:correlation_norm}, removing the normalization step does not significantly alter the ranking of generated outputs; the mean $\tau$ remains very high for all case studies and all models.

If we instead look at the actual distribution of the value and originality components (reported in Table \ref{covo_scores_unnorm}), we observe that removing the normalization can even double the originality score, while having a smaller impact on the value score. This is in line with our expectations, as discussed in Section \ref{method}.
Then, to evaluate the actual impact on training, we repeat all experiments from Section \ref{experiments} after removing the normalization step, i.e., by using the formula from Equation \ref{eq:length_normalized_score} as the reward.


\begin{table*}[ht]
\centering
\resizebox{\textwidth}{!}{%
\begin{tabular}{|l|cccc|cccc|}
 \hline
 & \multicolumn{4}{c|}{In-distribution} & \multicolumn{4}{c|}{Out-of-distribution} \\
 \cline{2-9}
 \multirow{-2}{*}{Method} & Corr. $\uparrow$ & Metric (L/S) $\uparrow$ & T-LCS $\downarrow$ & Tone $\uparrow$ & Corr. $\uparrow$ & Metric (L/S) $\uparrow$ & T-LCS $\downarrow$ & Tone $\uparrow$  \\
 \hline
 \texttt{Meta-Llama-3-8B} & $0.944$ & $0.360$ / $0.162$ & $9.640$ / $149$ & $0.634_{\pm 0.055}$ & $0.960$ & $0.413$ / $0.099$ & $6.288$ / $68$ & $0.610_{\pm 0.054}$ \\
 + CoVO                   & $0.928$ & $0.080$ / $0.061$ & $4.712$ / \textcolor{white}{00}$8$ & $0.662_{\pm 0.051}$ & $0.904$ & $0.240$ / $0.056$ & $4.704$ / $10$ & $0.637_{\pm 0.044}$ \\
 + CoVO (no-norm)         & $0.896$ & $0.300$ / $0.042$ & $4.336$ / \textcolor{white}{00}$6$ & $0.671_{\pm 0.046}$ & $0.912$ & $0.107$ / $0.018$ & $4.296$ / \textcolor{white}{0}$6$ & $0.643_{\pm 0.041}$ \\
 + CoVO + KL              & $0.936$ & $0.300$ / $0.132$ & $5.072$ / \textcolor{white}{0}$16$ & $0.661_{\pm 0.049}$ & $0.992$ & $0.347$ / $0.113$ & $4.992$ / \textcolor{white}{0}$9$ & $0.667_{\pm 0.047}$ \\
 + CoVO (no-norm) + KL    & $0.968$ & $0.300$ / $0.167$ & $5.128$ / \textcolor{white}{0}$16$ & $0.682_{\pm 0.049}$ & $0.968$ & $0.347$ / $0.085$ & $4.968$ / \textcolor{white}{0}$7$ & $0.665_{\pm 0.048}$ \\
 \hline
 \texttt{Qwen-2.5-3B}     & $0.992$ & $0.520$ / $0.102$ & $5.112$ / \textcolor{white}{00}$8$ & $0.817_{\pm 0.041}$ & $0.976$ & $0.600$ / $0.096$ & $4.856$ / \textcolor{white}{0}$8$ & $0.732_{\pm 0.049}$ \\
 + CoVO                   & $0.960$ & $0.440$ / $0.126$ & $5.032$ / \textcolor{white}{00}$7$ & $0.744_{\pm 0.051}$ & $0.984$ & $0.493$ / $0.092$ & $4.888$ / \textcolor{white}{0}$7$ & $0.694_{\pm 0.050}$ \\
 + CoVO (no-norm)         & $0.984$ & $0.420$ / $0.095$ & $4.936$ / \textcolor{white}{00}$7$ & $0.755_{\pm 0.051}$ & $0.952$ & $0.520$ / $0.089$ & $4.840$ / \textcolor{white}{0}$7$ & $0.698_{\pm 0.05}$ \\
 + CoVO + KL              & $0.984$ & $0.440$ / $0.115$ & $5.056$ / \textcolor{white}{00}$7$ & $0.779_{\pm 0.048}$ & $0.984$ & $0.507$ / $0.093$ & $4.840$ / \textcolor{white}{0}$8$ & $0.690_{\pm 0.053}$ \\
 + CoVO (no-norm) + KL    & $0.984$ & $0.460$ / $0.106$ & $5.064$ / \textcolor{white}{00}$7$ & $0.752_{\pm 0.052}$ & $0.960$ & $0.547$ / $0.092$ & $4.768$ / \textcolor{white}{0}$7$ & $0.714_{\pm 0.050}$ \\
 \hline
\end{tabular}
}
\caption{Aggregate results over 5 seeds of generated poems from training prompts (left) and test prompts (right), with and without the max-probability normalization step. Scores for poetic metrics are reported at the line (L) and syllable (S) levels and only consider requests for styles with specific metrical properties. Under T-LCS, we report both the mean and maximum longest common substring across all generated poems. The mean and the 95\% confidence interval are reported for tone adherence.\label{poetry_generation_unnorm}}
\end{table*}

\begin{figure}[ht]
    \centering
    \includegraphics[width=0.48\textwidth]{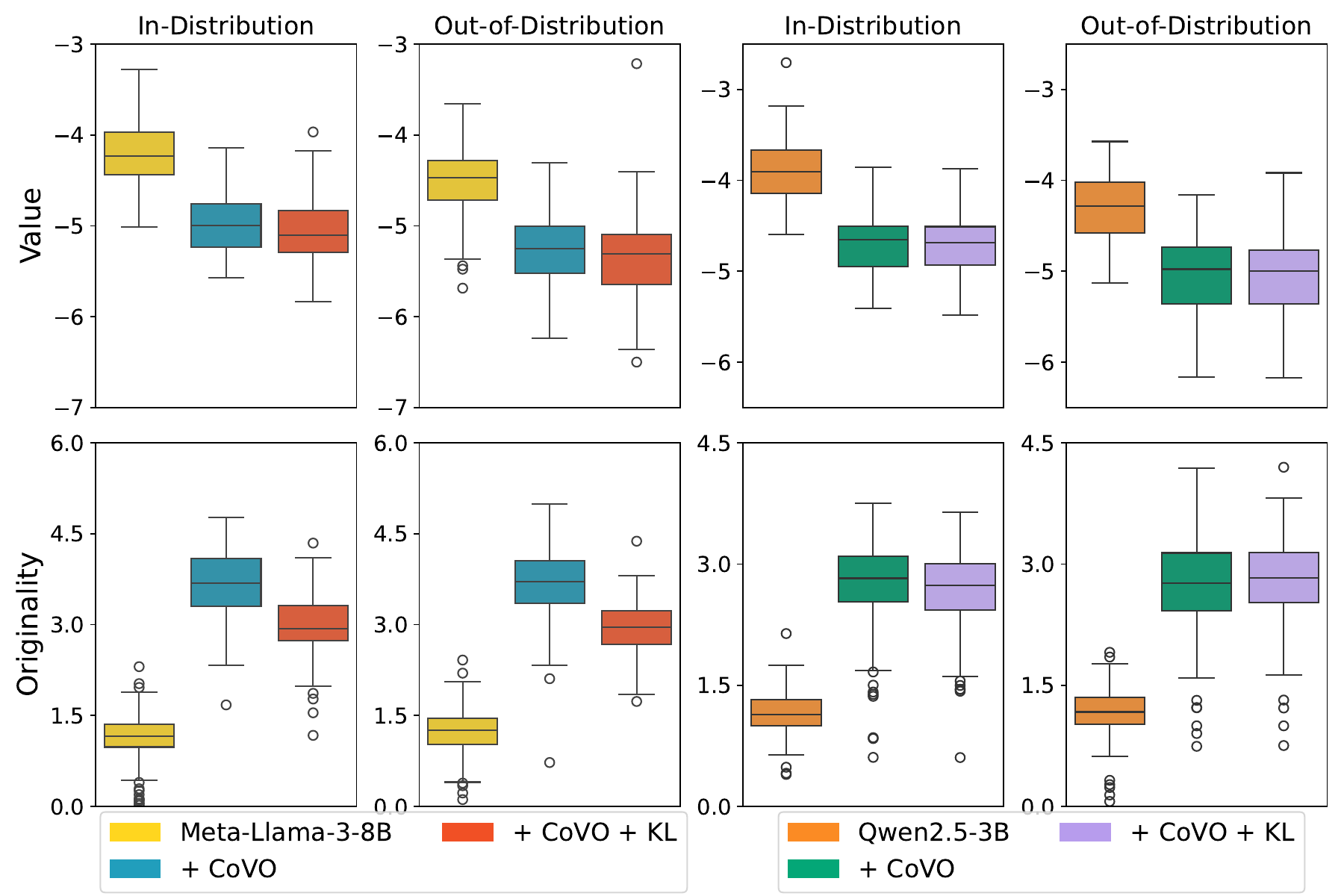}
    \caption{The distribution of value and originality scores for the in-distribution and out-of-distribution poems generated by the original models and our tuned versions without the normalization step.}
    \label{fig:val_orig_unnorm}
\end{figure}

For poetry generation, the effect of removing the normalization step is very subtle and has a minimal impact on aggregate scores. As reported in Table \ref{poetry_generation_unnorm}, the main effects occur on out-of-distribution prompts, where both correctness and poetic metric adherence tend to decrease, despite all other metrics being substantially similar to those obtained with the normalization step. However, upon closer inspection of the outputs produced by the models\footnote{All generated data can be found, together with code, at: \url{https://anonymous.4open.science/r/CoVO-grpo/}}, generated poems from models tuned without the normalization step tend to be more ungrammatical, with multiple repetitions, while still preserving the line-based structure of the text (which explains the preservation of line-based metric adherence) and a general meaning close to the requested tone (which explains the non-substantial changes in tone adherence). This is due to the unbalanced nature of the non-normalized CoVO components: as can be seen in Figure \ref{fig:val_orig_unnorm}, the unnormalized originality dominates the unnormalized value, and the KL penalty only mitigates its impact, making the overall optimization a mere deviation from the original model (and from the linguistic rules it encodes).

\begin{table*}[ht]
\centering
\resizebox{\textwidth}{!}{%
\begin{tabular}{|L{5.0cm}||C{1.9cm}|C{1.9cm}|C{1.4cm}C{1.4cm}||C{2.1cm}|C{1.9cm}|C{1.4cm}C{1.4cm}|} 
\hline
& \multicolumn{4}{c||}{GSM8K} & \multicolumn{4}{c|}{MATH} \\
\cline{2-9}
& & & \multicolumn{2}{c||}{Cross-Input Diversity} & & & \multicolumn{2}{c|}{Cross-Input Diversity} \\
\multirow{-3}{*}{\parbox{3cm}{\texttt{MetaMath-\\Mistral-7B}}} & \multirow{-2}{*}{Accuracy} & \multirow{-2}{*}{Entropy} & EAD & SBERT & \multirow{-2}{*}{Accuracy} & \multirow{-2}{*}{Entropy} & EAD & SBERT \\
\hline
Original model              & $77.96\% \;(3)$ & $0.156_{\pm .002}$ & $2.0071$ & $0.6402$ & $33.55\% \;(483)$ & $0.185_{\pm .002}$ & $5.7239$ & $\mathbf{0.8032}$ \\
+ Ext                       & $78.18\% \;(4)$ & $0.154_{\pm .002}$ & $2.0045$ & $\mathbf{0.6404}$ & $33.19\% \;(469)$ & $0.182_{\pm .002}$ & $5.7187$ & $0.8027$ \\
+ Ext + KL                  & $78.08\% \;(5)$ & $0.154_{\pm .002}$ & $2.0077$ & $0.6401$ & $33.56\% \;(476)$ & $0.182_{\pm .002}$ & $5.7517$ & $0.8028$ \\
+ CoVO                      & $78.12\% \;(3)$ & $\mathbf{0.164_{\pm .003}}$ & $2.0509$ & $0.6403$ & $33.29\% \;(521)$ & $\mathbf{0.197_{\pm .002}}$ & $5.8219$ & $0.8029$ \\
+ CoVO (no-norm)            & $77.88\% \;(3)$ & $\mathbf{0.164_{\pm .003}}$ & $\mathbf{2.0747}$ & $0.6400$ & $33.49\% \;(554)$ & $\mathbf{0.196_{\pm .002}}$ & $\mathbf{5.9046}$ & $0.8025$ \\
+ CoVO + KL                 & $78.12\% \;(3)$ & $\mathbf{0.163_{\pm .003}}$ & $2.0464$ & $0.6402$ & $32.79\% \;(533)$ & $\mathbf{0.195_{\pm .002}}$ & $5.8442$ & $0.8030$ \\
+ CoVO (no-norm) + KL       & $77.37\% \;(3)$ & $\mathbf{0.163_{\pm .003}}$ & $2.0647$ & $0.6400$ & $33.00\% \;(564)$ & $\mathbf{0.195_{\pm .002}}$ & $5.8926$ & $0.8028$ \\
+ CoVO + Ext                & $78.33\% (4)$ & $0.158_{\pm .002}$ & $2.0340$ & $\mathbf{0.6404}$ & $\mathbf{33.76}\% (503)$ & $0.188_{\pm .002}$ & $5.8114$ & $0.8030$ \\
+ CoVO (no-norm) + Ext      & $78.19\% (3)$ & $0.157_{\pm .002}$ & $2.0441$ & $0.6397$ & $33.24\% (506)$ & $0.187_{\pm .002}$ & $5.8413$ & $0.8024$ \\
+ CoVO + Ext + KL           & $77.95\% \;(4)$ & $0.157_{\pm .002}$ & $2.0367$ & $0.6402$ & $33.74\% \;(497)$ & $0.187_{\pm .002}$ & $5.8082$ & $0.8031$ \\
+ CoVO (no-norm) + Ext + KL & $\mathbf{78.55}\% \;(4)$ & $0.157_{\pm .002}$ & $2.0416$ & $0.6399$ & $33.61\% \;(515)$ & $0.187_{\pm .002}$ & $5.8133$ & $0.8029$ \\
\hline
\end{tabular}
}
\caption{Accuracy and diversity of results for the GSM8K (left) and MATH (right) test sets with and without the max-probability normalization step. In brackets, the number of responses exceeding the maximum token limit. The mean and the 95\% confidence interval are reported for sequence-level entropy.
\label{math_covo_unnorm}}
\end{table*}

Table \ref{math_covo_unnorm} reports the results for the math problem-solving case study with and without the normalization step. As can be noticed, while unnormalized scores lead to an increase in cross-input EAD, i.e., over a wider selection of different tokens, the entropy (as well as the cross-input SBERT score) is not higher, but rather slightly decreased. This suggests that optimizing for the unnormalized score tends to modify the greedy selection, whereas the normalized score tends to preserve it more, while also flattening the distribution more. While the difference is subtle, it has a direct impact on quality: whenever the original model already performs well on the task at hand, changing the greedy selection, i.e., using the unnormalized score, can negatively affect accuracy (as happens for GSM8K); on the other hand, if the original model struggles on the task at hand, simply flattening the distribution is less useful (as suggested by the higher accuracy of unnormalized scores for MATH).

\begin{table}[ht]
\centering
\resizebox{.49\textwidth}{!}{%
\begin{tabular}{|l|cc|}
 \hline
 Method & Quality & Novelty \\
 \hline
 \texttt{Llama3.2-3B-Instruct} & $3.90 \pm 0.16$ & $4.31 \pm 0.23$ \\
 + CoVO                        & $3.89 \pm 0.20$ & $\mathbf{8.71 \pm 0.16}$ \\
 + CoVO (no-norm)              & $3.80 \pm 0.20$ & $\mathbf{8.41 \pm 0.17}$ \\
 + CoVO + KL                   & $\mathbf{4.88 \pm 0.19}$ & $7.13 \pm 0.23$ \\
 + CoVO (no-norm) + KL         & $\mathbf{4.62 \pm 0.19}$ & $7.07 \pm 0.24$ \\
 \hline
 \texttt{Qwen-2.5-7B-Instruct} & $3.51 \pm 0.16$ & $3.62 \pm 0.23$ \\
 + CoVO                        & $\mathbf{3.91 \pm 0.18}$ & $4.52 \pm 0.24$ \\
 + CoVO (no-norm)              & $\mathbf{3.99 \pm 0.18}$ & $\mathbf{5.73 \pm 0.25}$ \\
 + CoVO + KL                   & $\mathbf{3.85 \pm 0.17}$ & $4.31 \pm 0.23$ \\
 + CoVO (no-norm) + KL         & $\mathbf{3.73 \pm 0.17}$ & $4.03 \pm 0.23$ \\
 \hline
\end{tabular}
}
\caption{The utility (quality) and distinct (novelty) scores on NoveltyBench (curated partition) for the original models and our methods (tuned on the wildchat partition) with and without the max-probability normalization step. Mean and 95\% confidence intervals are reported over 5 seeds.
\label{noveltybench_table_unnorm}}
\end{table}

Finally, Table \ref{noveltybench_table_unnorm} reports the results for the NoveltyBench case study. Here, the main effect of normalization is preserving the value score, keeping the model closer to what the original model recognizes as effective. This means that, for an already well-performing model such as \texttt{Llama3.2-3B-Instruct}, removing the normalization reduces the final scores. On the other hand, for a less-performing model such as \texttt{Qwen-2.5-7B-Instruct}, larger deviations from the original model and reduced importance of the value score do not lower the scores, making normalization not significantly better (when including the KL penalty) or even detrimental (when optimizing for the CoVO score alone).
In summary, the normalization plays a key role in automatically balancing the value and originality components of the CoVO score, which is generally desirable, although there are particular situations in which this balancing is less effective.
\section{Ablation Study on the Prompt Sensitivity for Value Score} \label{ablation_prompt}
The calculation of the value score, i.e., $p_\theta(\mathbf{z}|\mathbf{x})$, is based on the approximation $p_\theta(\mathbf{z}|\mathbf{x'})$ with $\mathbf{x'} = \mathbf{x} + \mathbf{q}$, where $\mathbf{q}$ represents a plain additional sentence or any other simple prompt-level trick designed to increase the likelihood of $\mathbf{z}$ (as well as of alternative but reasonable $\mathbf{\hat{z}}$). While this is necessary in practice, introducing an additional variable can affect the final CoVO score computation, and different choices of $\mathbf{q}$ values can lead to different behaviors. Therefore, we study the sensitivity to different $\mathbf{q}$ formulations and their impact on the use of the CoVO score as a reward in a GRPO setting. In particular, we define two additional $\mathbf{q}$ formulations for each of the three case studies, examining whether they lead to different output rankings and, consequently, different downstream performance.

For poetry generation, we adopt the following two alternative prompts:

{\small
\begin{tcolorbox}[colback=white,colframe=black]
What was the instruction leading to the following poem?\\
\\
\#\#\# Poem:\\
\{\texttt{poem}\}\\
\\
\#\#\# Instruction: Write a
\end{tcolorbox}
}

\noindent and

{\small
\begin{tcolorbox}[colback=white,colframe=black]
Try to guess what instruction led to the following poem:\\
\\
\{\texttt{poem}\}\\
\\
Instruction: Write a
\end{tcolorbox}
}

\noindent where the first is essentially a slightly modified version of the original prompt (i.e., with one additional word), while the second concerns a complete rephrasing.

Similarly, for math problem solving, we adopt the two alternative prompts reported below:

{\small
\begin{tcolorbox}[colback=white,colframe=black]
Below is a solution that appropriately solves a problem. Write the instruction that describes the problem.\\
\\
\#\#\# Solution:\\
\{\texttt{response}\}\\
\\
\#\#\# Instruction:
\end{tcolorbox}
}

\noindent and

{\small
\begin{tcolorbox}[colback=white,colframe=black]
Try to guess what instruction led to the following response.\\
\\
\#\#\# Response:\\
\{\texttt{response}\}\\
\\
\#\#\# Instruction:
\end{tcolorbox}
}

Finally, for NoveltyBench, we test with the following prompt:

{\small
\begin{tcolorbox}[colback=white,colframe=black]
Below is a response that appropriately addresses a request. Write the instruction that describes the request.\\
\\
\#\#\# Response:\\
\{\texttt{response}\}\\
\\
\#\#\# Instruction:
\end{tcolorbox}
}

\noindent and

{\small
\begin{tcolorbox}[colback=white,colframe=black]
Try to guess what instruction led to the following response.\\
\\
\#\#\# Response:\\
\{\texttt{response}\}\\
\\
\#\#\# Instruction:
\end{tcolorbox}
}

\begin{figure*}[ht]
    \centering
    \includegraphics[width=0.98\textwidth]{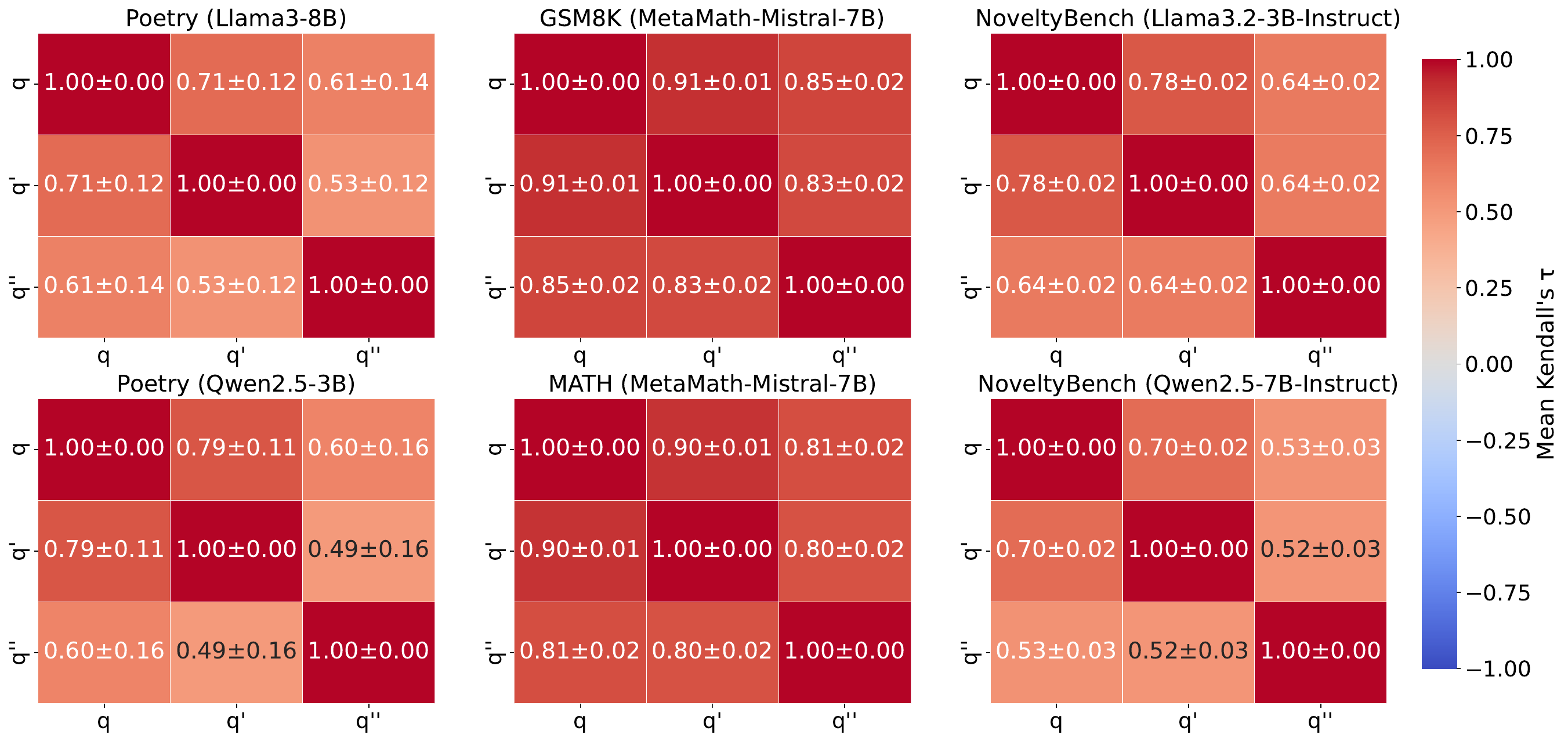}
    \caption{Correlation matrices between the CoVO value component for our three case studies with different $q$ prompts. Kendall's $\tau$ mean and 95\% confidence interval between scores for multiple solutions over the same inputs are reported.}
    \label{fig:correlation_q}
\end{figure*}

First, we evaluate whether the value component obtained with one prompt correlates with the value component obtained with a different prompt when scoring and ranking $G=4$ different outputs returned for the same prompt. In this way, we can evaluate the actual impact of different prompts when using the score as a reward. Figure \ref{fig:correlation_q} reports the correlation matrices for each case study across the three different prompts, where each correlation represents the mean and 95\% confidence interval of Kendall's $\tau$ correlation across multiple inputs (all 25 possible training prompts for poetry generation; a sample of 1,000 training prompts for GSM8K and MATH; and all inputs from the NoveltyBench \texttt{wildchat} partition). In general, the alternative prompt formulations have a strong correlation with the selected one across all case studies and selected models. Math problem solving shows the strongest correlation for both $q'$ and $q''$; conversely, both poetry generation and NoveltyBench have slightly lower values, even if correlation remains significantly positive.

\begin{table}[ht]
\centering
\resizebox{.48\textwidth}{!}{%
\begin{tabular}{|L{3.8cm}|C{1.5cm}C{3.1cm}C{3.1cm}|}
 \hline
 Experiment & $q$ & $q'$ & $q''$ \\
 \hline
 Poetry generation & & & \\
 \hline
 \texttt{Llama3-8B}   & $-4.29_{\pm 0.06}$ & $-4.07_{\pm 0.06}\;(5.2\%)$ & $-4.18_{\pm 0.06}\;(2.8\%)$ \\
 \texttt{Qwen-2.5-3B} & $-3.82_{\pm 0.06}$ & $-3.80_{\pm 0.06}\;(1.4\%)$ & $-3.99_{\pm 0.07}\;(5.1\%)$ \\
 \hline
 Math problem solving & & & \\
 \hline
 GSM8K & $-1.17_{\pm 0.01}$ & $-1.19_{\pm 0.01}\;(2.5\%)$ & $-1.14_{\pm 0.01}\;(3.8\%)$ \\
 MATH  & $-0.91_{\pm 0.01}$ & $-0.94_{\pm 0.01}\;(4.4\%)$ & $-0.91_{\pm 0.01}\;(6.6\%)$ \\
 \hline
 NoveltyBench & & & \\
 \hline
 \texttt{Llama3.2-3B-Instruct} & $-1.88_{\pm 0.02}$ & $-1.74_{\pm 0.02}\;(8.8\%)$ & $-1.91_{\pm 0.02}\;(7.8\%)$ \\
 \texttt{Qwen-2.5-7B-Instruct} & $-2.87_{\pm 0.04}$ & $-2.95_{\pm 0.04}\;(7.0\%)$ & $-2.26_{\pm 0.03}\;(20.8\%)$ \\
 \hline
\end{tabular}
}
\caption{The value score at varying different $q$ prompts. Mean $\pm$ 95\% confidence intervals (and percentage mean absolute variance with respect to the original prompt formulation) are reported over 4 generated outputs from the entire training set for poetry and NoveltyBench and from a sample of 1000 training data for GSM8K and MATH.
\label{value_scores_different_prompts}}
\end{table}

However, considering the effect of different $q$ prompts by looking at the value component in isolation is not sufficient. The proposed framework optimizes for the sum of the value and originality components; thus, the magnitude of the value score for different $q$ is equally important, as it impacts the trade-off between effectiveness and novelty. As reported in Table \ref{value_scores_different_prompts}, the variations are very small: the value scores are very close to each other regardless of what $q$ prompt we considered, and only for NoveltyBench do we observe a few potentially significant differences.

\begin{table}[ht]
\centering
\resizebox{.48\textwidth}{!}{%
\begin{tabular}{|l|cc|}
 \hline
 Method & Quality & Novelty \\
 \hline
 \texttt{Llama3.2-3B-Instruct} & $3.90 \pm 0.16$ & $4.31 \pm 0.23$ \\
 + CoVO ($q$)                  & $3.89 \pm 0.20$ & $\mathbf{8.71 \pm 0.16}$ \\
 + CoVO ($q'$)                 & $4.16 \pm 0.21$ & $\mathbf{8.45 \pm 0.18}$ \\
 + CoVO ($q''$)                & $4.00 \pm 0.21$ & $\mathbf{8.50 \pm 0.18}$ \\
 + CoVO ($q$) + KL             & $\mathbf{4.88 \pm 0.19}$ & $7.13 \pm 0.23$ \\
 + CoVO ($q'$) + KL            & $\mathbf{4.71 \pm 0.19}$ & $6.37 \pm 0.25$ \\
 + CoVO ($q''$) + KL           & $\mathbf{5.00 \pm 0.19}$ & $6.89 \pm 0.23$ \\
 \hline
 \texttt{Qwen-2.5-7B-Instruct} & $3.51 \pm 0.16$ & $3.62 \pm 0.23$ \\
 + CoVO ($q$)                  & $3.91 \pm 0.18$ & $4.52 \pm 0.24$ \\
 + CoVO ($q'$)                 & $3.81 \pm 0.17$ & $4.26 \pm 0.24$ \\
 + CoVO ($q''$)                & $\mathbf{4.50 \pm 0.18}$ & $\mathbf{6.03 \pm 0.25}$ \\
 + CoVO ($q$) + KL             & $3.85 \pm 0.17$ & $4.31 \pm 0.23$ \\
 + CoVO ($q'$) + KL            & $3.62 \pm 0.17$ & $3.88 \pm 0.23$ \\
 + CoVO ($q''$) + KL           & $3.99 \pm 0.17$ & $4.71 \pm 0.24$ \\
 \hline
\end{tabular}
}
\caption{The utility (quality) and distinct (novelty) scores on NoveltyBench (curated partition) for the original models and our methods (tuned on the wildchat partition) by varying $q$ prompt formulation. Mean and 95\% confidence intervals are reported over 5 seeds.
\label{noveltybench_table_q}}
\end{table}

Given that NoveltyBench is the only case study where both correlation and mean variance suggest that different $q$ prompts may elicit different behaviors when optimizing for the CoVO score, we finally conduct an ablation study by tuning both base models to maximize the CoVO score with the $q'$ and $q''$ prompt formulations. As reported in Table \ref{noveltybench_table_q}, different prompt formulations can sometimes elicit different behaviors, even if all tested formulations achieve significantly better performance than the original models. In the case of \texttt{Llama3.2-3B-Instruct}, different formulations lead to non-significantly different scores, and they only slightly modify the quality-novelty trade-off. As for \texttt{Qwen-2.5-7B-Instruct}, the closer formulation $q'$ does not significantly differ from the original $q$ formulation, even if it tends to achieve lower mean results; however, the $q''$ formulation achieves consistently higher scores on both quality and novelty metrics. Overall, these results suggest that, even if different $q$ prompt formulations can lead to non-significant mathematical differences, the $q$ prompt formulation is, in all respects, a relevant hyperparameter, and careful tuning can lead to improved results.

\section{GutenVerse Dataset} \label{gutenverse_dataset}
To evaluate the accidental reproduction rate of generated poems, we introduce the GutenVerse dataset, which comprises over 84,000 public-domain, English-written poems extracted from Project Gutenberg. While generated poems can reproduce different content, e.g., songs or copyrighted material, we believe this can provide a useful indication of how likely a text is to be original or not.

To derive our dataset, we started with \textit{Gutenberg, dammit}\footnote{\url{https://github.com/aparrish/gutenberg-dammit/}}, a corpus of all plaintext files in Project Gutenberg (up to June 2016). We selected all text files whose metadata report English as the language, public domain as the copyright status, \textit{poetry} among the subjects, or \textit{poems} or \textit{poetical work} in the title, and that were not translations of another book. Then, we applied a series of rules (e.g., regarding verse length) to automatically extract titles and poems from all selected text files, thereby defining our GutenVerse dataset. While it can still contain content that is not poetry (e.g., a table of contents formatted very uncommonly), the poems can be effectively used to measure the overlap between real and generated text.

We also released a datasheet \cite{gebru2018datasheets} for the GutenVerse dataset that can be found, together with the code used to create it, at: \url{https://anonymous.4open.science/r/GutenVerse-DD32/}.

\section{Qualitative Examples}

In the following sections, we discuss the generated outputs for all our case studies from a qualitative perspective. While we present only a few particularly significant examples here, all outputs, together with their relative scores, are available at: \url{https://anonymous.4open.science/r/CoVO-grpo}.

\subsection{Detailed Analysis of the Generated Poems} \label{poem_analysis}

We now present some noteworthy generated poems to provide a detailed qualitative discussion of our methods and our score.

The baseline model \texttt{Meta-Llama-3-8B}, while occasionally producing prose (i.e., without line breaks), is the most conservative method. This results in its poems being usually well-formatted but also somewhat \textit{banal}, and more prone to reproducing existing works. For example, when asked to write a dark hymn, it produced the very famous two lines ``I am the master of my fate: / I am the captain of my soul'' from \textit{Invictus} by William Ernest Henley. Similarly, when asked to write a happy sonnet, it produced \textit{My Shadow} by Robert Louis Stevenson; notably, the latter achieved both the lowest originality component and the highest T-LCS score. This strongly suggests that the model has memorized it and reproduced it almost entirely, with only minor alterations in punctuation (enough to prevent the LCS score from being even higher). We report them in Table \ref{highest_lcs_baseline}. The originality score also plays a key role in the overall score: the aforementioned poem is the one with the lowest CoVO score, while the poem achieving the highest CoVO score also achieves the highest originality score (see Table \ref{best_worst_baseline}).

Regarding the fine-tuned models, the one trained without the KL loss exhibits the opposite behavior. It never reproduces existing poems but occasionally exploits the CoVO score in adversarial ways, for example, by learning to generate specific words associated with the requested tone. In one instance, it even learned to simply write ``a poem'', which resulted in one of the highest possible scores. Conversely, maximizing the originality component leads the model to produce noteworthy outputs, but not always ones that satisfy the required stylistic constraints, as reported in Table \ref{best_orig_covo}, and sometimes even ungrammatical ones. In general, optimizing the CoVO score without the KL loss leads the model to deviate from memorized poems, but also from metrical rules, and leads it to focus on the tone (both the explicit, requested tone and that intrinsic to the poetic style). For example, its limericks maintain the typical playful tone, but only once out of 25 poems (5 for each seed) adheres to the traditional ``There [once] was a man'' starting line and only partially, as it begins with ``There was an old dog''; moreover, this is not even the highest-valued limerick.

By contrast, including the KL loss makes the model adhere more closely to the required style; for instance, many more limericks begin with variations of the classic ``There [once] was a man'', but this also holds for haikus and out-of-distribution requests (see Table \ref{best_covo_kl}). In general, the KL loss seems to help the model preserve coherence and improve the CoVO score in a more meaningful way, e.g., the repetitions are quite rare (though, when present, they again lead to the highest value scores), as well as the regurgitation of existing poems.

The qualitative findings for the \texttt{Qwen2.5-3} model and its CoVO- and CoVO+KL-tuned versions are quite similar, even though they are influenced by the different capabilities of the base model. Indeed, it never produces prosaic text, and its poems are better formed. However, it tends to repeat the same sentences multiple times, sometimes even in a meaningless way (e.g., by always repeating the same line). However, these repetitions can also lead to valuable and original outputs, as exemplified in Table \ref{examples_qwen_baseline}.

Finally, the \texttt{Qwen2.5-3} models tuned with the CoVO score (alone or with KL) behave similarly to those reported above. In particular, the highest-scoring poems do not contain repetitions and fulfill all poetic constraints (see Table \ref{best_qwen_tuned}), and they exploit the score in an adversarial way only very rarely. However, when this happens, it can occasionally be incorporated positively. For example, both the model tuned with the CoVO score alone and the model tuned with CoVO+KL produced two limericks that, despite containing several repetitions of the same items, are still meaningful and correctly capture the general spirit of limericks (see Table \ref{qwen_covo_limericks}).

\subsection{Math Problem Solving Output Exemplars}

We conducted a qualitative analysis of the solutions provided by our methods and the baselines for the two math datasets. As also summarized by the quantitative metrics reported in the main paper, we found that the use of the CoVO score, especially without KL, moves the model away from its most probable solutions, while the adoption of the external reward alone (with or without KL) is often not sufficient to alter the most probable answer in a meaningful way. However, the resulting solutions are highly similar across all cases. This suggests that the effect of the CoVO score is to redistribute the model’s probability mass away from the originally most probable solutions and toward closely related, high-probability alternatives, in which only specific parts of the chain of thought are modified. While this can also impact the final numerical answer (as reported in the example from Table \ref{gsm8k_example_1}), it often means that the output has switched between alternative paths toward the same final response (as in Table \ref{gsm8k_example_2}).

\subsection{NoveltyBench Output Analysis}

A careful analysis of the generated outputs and the relative scores from NoveltyBench reveals that the impact of the CoVO score is particularly evident at the form level. Indeed, the answers to the questions are often semantically similar, sharing the same objects, but the sentences are formed differently: wherever the baseline would follow one or two templates, our training pushes the model to adopt multiple templates. This happens for both models we tested. As far as the \texttt{Llama3.2-3B-Instruct} model is concerned, Table \ref{llama_nb_examples1} reports two examples of this. However, sometimes the CoVO score helps the model output a greater variety of semantic options as well, covering more exemplars of the same element or diverging from the most common terms (as in the second example from Table \ref{llama_nb_examples2}, where our methods produce more titles without banal terms such as `translation', `bits', and their variants).

The same considerations also apply to \texttt{Qwen2.5-7B-Instruct}, where these effects are sometimes even more apparent due to its lower output diversity, which often leads to identical or only slightly varied responses. 
The effect of the CoVO score is to yield outputs that follow alternative templates, but are sometimes also characterized by distinct items or themes. Tables \ref{qwen_nb_example_1} and \ref{qwen_nb_example_2} report two examples that clearly manifest these features: in the first case, all the baseline outputs are almost identical, while our methods produce outputs that share similar templates (potentially slightly modified) but with two or three different content; in the second case, all the baseline outputs are the same, while our methods produce outputs that diverge both in terms of semantic content and in terms of used templates.
%
%
%
%
\begin{table}[t]
\centering
\resizebox{0.48\textwidth}{!}{%
\begin{tabular}{|L{4.2cm}|L{3.3cm}|} 
 \hline
    \multicolumn{2}{|c|}{Dark Hymn}\\
    \hline
    \multicolumn{2}{|l|}{\specialcell[l]{I am the master of my fate:\\I am the captain of my soul.}}\\
    \hline
    $V$ = -4.495 & $O$ = 0.259\\
    \hline
 \multicolumn{2}{|c|}{Happy sonnet} \\
 \hline
 \multicolumn{2}{|l|}{\specialcell[l]{i have a little shadow that goes in and out with me,\\and what can be the use of him is more than i can see.\\he is very, very like me from the heels up to the head;\\and i see him jump before me, when i jump into my bed.\\the funniest thing about him is the way he likes to grow—\\not at all like proper children, which is always very slow;\\for he sometimes shoots up taller like an india-rubber ball,\\and he sometimes gets so little that there's none of him at all.\\he hasn't got a notion of how children ought to play,\\and can only make a fool of me in every sort of way.\\he stays so close beside me, he's a coward you can see;\\i'd think shame to stick to nursie as that shadow sticks to me!\\one morning, very early, before the sun was up,\\i rose and found the shining dew on every buttercup;\\but my lazy little shadow, like an arrant sleepy-head,\\had stayed at home behind me and was fast asleep in bed.}} \\
 \hline
 $V$ = -4.485 & $O$ = 0.017 \\
 \hline
\end{tabular}
}
\caption{The poems generated by the \texttt{Meta-Llama-3-8B} model with the highest T-LCS score and the lowest originality score.\label{highest_lcs_baseline}}
\end{table}
\begin{table}[t]
\centering
\small
\resizebox{0.48\textwidth}{!}{%
\begin{tabular}{|L{3.3cm}|L{3.3cm}|} 
    \hline
    \multicolumn{2}{|c|}{Poignant tanka}\\
    \hline
    \multicolumn{2}{|l|}{\specialcell[l]{if your dreams\\cannot comfort you,\\why not try reality…?}}\\
    \hline
    $V$ = -4.280 & $O$ = 2.416\\
    \hline
\end{tabular}
}
\caption{The poem generated by the \texttt{Meta-Llama-3-8B} model with the highest overall CoVO score, which is also the one with the highest originality score.\label{best_worst_baseline}}
\end{table}
\begin{table}[t]
\centering
\small
\resizebox{0.48\textwidth}{!}{%
\begin{tabular}{|L{3.3cm}|L{3.3cm}|} 
    \hline
    \multicolumn{2}{|c|}{Dark ballad}\\
    \hline
    \multicolumn{2}{|l|}{\specialcell[l]{a little silver is\\an awful way\\a few\\and it's all.}}\\
    \hline
    $V$ = -3.532 & $O$ = 2.150\\
    \hline
    \multicolumn{2}{|c|}{Nostalgic couplet}\\
    \hline
    \multicolumn{2}{|l|}{\specialcell[l]{A spondaic ha ha\\to be sibilant of our first love\\to be, a small gawking of an old songless.}}\\
    \hline
    $V$ = -3.443 & $O$ = 1.814\\
    \hline
\end{tabular}
}
\caption{The poems generated by the CoVO-based fine-tuned \texttt{Meta-Llama-3-8B} model with the highest originality scores from training (top) and testing (bottom) tone-style pairs.\label{best_orig_covo}}
\end{table}
%
%
%
%
\begin{table}[t]
\centering
\small
\resizebox{0.48\textwidth}{!}{%
\begin{tabular}{|L{3.3cm}|L{3.3cm}|} 
    \hline
    \multicolumn{2}{|c|}{Dark haiku}\\
    \hline
    \multicolumn{2}{|l|}{\specialcell[l]{Ceremony: white hair\\holy woman, my shadow:\\a leaf, then light}}\\
    \hline
    $V$ = -3.877 & $O$ = 1.908\\
    \hline
    \multicolumn{2}{|c|}{Solemn couplet}\\
    \hline
    \multicolumn{2}{|l|}{\specialcell[l]{What is so rare and fragile,\\The world can crush with only a kiss.}}\\
    \hline
    $V$ = -4.083 & $O$ = 1.957\\
    \hline
\end{tabular}
}
\caption{Two high-scoring, structure-preserving poems from the CoVO+KL-based fine-tuned \texttt{Meta-Llama-3-8B} model.\label{best_covo_kl}}
\end{table}
\begin{table}[t]
\centering
\small
\resizebox{0.48\textwidth}{!}{%
\begin{tabular}{|L{3.3cm}|L{3.3cm}|} 
    \hline
    \multicolumn{2}{|c|}{Romantic hymn}\\
    \hline
    \multicolumn{2}{|l|}{\specialcell[l]{how can i fail to love thee\\when thou are thou\\how can i forget such magic\\when eyes are blue\\how can i resist the call of thy\\when heart is my truest desire\\how shall i ever live without thee\\when thou art mine.}}\\
    \hline
    $V$ = -1.732 & $O$ = 1.224\\
    \hline
    \multicolumn{2}{|c|}{Mysterious hymn}\\
    \hline
    \multicolumn{2}{|l|}{\specialcell[l]{all the things they said to them,\\they saw,\\but to them had no meaning.}}\\
    \hline
     $V$ = -4.597 & $O$ = 2.140 \\
    \hline
\end{tabular}
}
\caption{The poems generated by the \texttt{Qwen2.5-3} model with the highest value score (top) and highest originality score (bottom).\label{examples_qwen_baseline}}
\end{table}
\begin{table}[t]
\centering
\small
\resizebox{0.48\textwidth}{!}{%
\begin{tabular}{|L{3.3cm}|L{3.3cm}|} 
    \hline
    \multicolumn{2}{|c|}{Romantic haiku}\\
    \hline
    \multicolumn{2}{|l|}{\specialcell[l]{blue evening cloaks\\a silver-white bride\\flutters in the night}}\\
    \hline
    $V$ = -2.784 & $O$ = 1.770\\
    \hline
    \multicolumn{2}{|c|}{Dark limerick}\\
    \hline
    \multicolumn{2}{|l|}{\specialcell[l]{the night was dim.\\darkness held its drum beat, a\\beat too long to keep.\\yet with a shudding crash, it fell\\onto a bed where a baby would steal it.}}\\
    \hline
     $V$ = -3.260 & $O$ = 1.922 \\
    \hline
\end{tabular}
}
\caption{The two highest-scoring poems generated by the tuned versions of the \texttt{Qwen2.5-3} model (top, tuned with CoVO alone; bottom, tuned with CoVO+KL).\label{best_qwen_tuned}}
\end{table}
\begin{table}[t]
\centering
\small
\resizebox{0.48\textwidth}{!}{%
\begin{tabular}{|L{3.3cm}|L{3.3cm}|} 
    \hline
    \multicolumn{2}{|c|}{Romantic limerick}\\
    \hline
    \multicolumn{2}{|l|}{\specialcell[l]{A love that’s not\\A love that’s not\\A love that’s not for the birds\\Is not\\A love, what birds make.\\A love is for the human heart, true, l’lady.}}\\
    \hline
    $V$ = -3.002 & $O$ = 1.222\\
    \hline
    \multicolumn{2}{|c|}{Mysterious limerick}\\
    \hline
    \multicolumn{2}{|l|}{\specialcell[l]{A witch's brew,\\As they sought to brew brew.\\
Witch's brew, brew,\\To their brew their brain brewed a dream.\\Witch's brew, brew.}}\\
    \hline
     $V$ = -3.644 & $O$ = 1.217 \\
    \hline
\end{tabular}
}
\caption{Two examples of how the tuned versions of the \texttt{Qwen2.5-3} model (top, tuned with CoVO alone; bottom, tuned with CoVO+KL) exploit repetitions to write meaningful poems.\label{qwen_covo_limericks}}
\end{table}

%
%
%
\begin{table*}[t]
\centering
\resizebox{\textwidth}{!}{%
\begin{tabular}{|m{3.5cm}|m{11cm}|}
\hline
\rowcolor{gray!20} & Joe has \$50 to buy an outfit for his new field trip. There is a 30\% off sale at the clothing store. The shirt he picks out has a price of \$25. He also picks out a pair of shorts for \$35. Assuming that sales tax is included, how much money will Joe have left after the purchase?\\
\cline{2-2}
\rowcolor{gray!20} \multirow{-5}{*}{Question} & Answer: 8 \\
\hline
\texttt{MetaMath-Mistral-7} & First, Joe has \$50 to spend.
The shirt is 30\% off, so it is discounted by 30/100 * \$25 = \$7.50
The discounted price of the shirt is \$25 - \$7.50 = \$17.50
The shorts are \$35.
The total cost of the outfit is \$17.50 + \$35 = \$52.50
Joe will have \$50 - \$52.50 = -\$2.50 left after the purchase.
Since Joe will have a negative amount left, it means he will not have any money left after the purchase.
\#\#\#\# 0\\
\hline
+ Ext. reward & First, Joe has \$50 to spend.
The shirt is 30\% off, so it is discounted by 30/100 * \$25 = \$7.50
The discounted price of the shirt is \$25 - \$7.50 = \$17.50
The shorts are \$35.
The total cost of the outfit is \$17.50 + \$35 = \$52.50
Joe will have \$50 - \$52.50 = -\$2.50 left after the purchase.
Since Joe will have a negative amount left, it means he will not have any money left after the purchase.
\#\#\#\# 0\\
\hline
+ Ext + KL & First, Joe has \$50 to spend.
The shirt is 30\% off, so it is discounted by 30/100 * \$25 = \$7.50
The discounted price of the shirt is \$25 - \$7.50 = \$17.50
The shorts are \$35.
The total cost of the outfit is \$17.50 + \$35 = \$52.50
Joe will have \$50 - \$52.50 = -\$2.50 left after the purchase.
Since Joe will have a negative amount left, it means he will not have any money left after the purchase.
\#\#\#\# 0\\
\hline
+ CoVO & First, Joe has \$50 to spend.
The shirt he picks out is \$25, so he will spend \$25 on the shirt.
The pair of shorts he picks out is \$35, so he will spend \$35 on the shorts.
The total amount he will spend is \$25 + \$35 = \$60.
Since there is a 30\% off sale, Joe will get 30\% off of \$60, which is \$60 * 0.30 = \$18 off.
So, the total amount he will pay after the discount is \$60 - \$18 = \$42.
Therefore, Joe will have \$50 - \$42 = \$8 left after the purchase.
\#\#\#\# 8\\
\hline
+ CoVO + KL & Joe has \$50 to spend.
There is a 30\% off sale, so he will pay 100\% - 30\% = 70\% of the original price.
The shirt is \$25, so he will pay 70\% of \$25 = \$17.50 for the shirt.
The shorts are \$35, so he will pay 70\% of \$35 = \$24.50 for the shorts.
The total cost of the outfit is \$17.50 + \$24.50 = \$42.
Joe will have \$50 - \$42 = \$8 left after the purchase.
\#\#\#\# 8\\
\hline
+ CoVO + Ext & First, Joe has \$50 to spend.
The shirt he picks out is \$25, so he will spend \$25 on the shirt.
The pair of shorts he picks out is \$35, so he will spend \$35 on the shorts.
The total amount he will spend is \$25 + \$35 = \$60.
Since there is a 30\% off sale, Joe will get 30\% off of \$60, which is \$60 * 0.30 = \$18 off.
So, the total amount he will pay after the discount is \$60 - \$18 = \$42.
Therefore, Joe will have \$50 - \$42 = \$8 left after the purchase.
\#\#\#\# 8\\
\hline
+ CoVO + Ext + KL & First, Joe has \$50 to spend.
The shirt he picks out is \$25, so that leaves him with \$50 - \$25 = \$25.
Next, he picks out a pair of shorts for \$35, so that leaves him with \$25 - \$35 = -\$10.
Since Joe has -\$10 left, it means he spent more than he had.
Therefore, Joe will have \$0 left after the purchase.
\#\#\#\# 0\\
\hline
\end{tabular}
}
\caption{\label{gsm8k_example_1} A qualitative example of GSM8K outputs for the base \texttt{MetaMath-Mistral-7} model and its tuned variants with different responses, both in terms of form and of content.}
\end{table*}

\begin{table*}
\centering
\small
\resizebox{\textwidth}{!}{%
\begin{tabular}{|m{3.5cm}|m{11cm}|}
\hline
\rowcolor{gray!20} & Jane has saved \$4 of her allowance every week for the past 8 weeks. If she continues to save the same amount each week, how many more weeks will it take for her to have saved a total of \$60?\\
\cline{2-2}
\rowcolor{gray!20} \multirow{-4}{*}{Question} & Answer: 7 \\
\hline
\texttt{MetaMath-Mistral-7} & Jane has saved \$4 every week for 8 weeks, so she has saved a total of \$4 x 8 = \$32.
She wants to save a total of \$60, so she still needs to save \$60 - \$32 = \$28.
If she continues to save \$4 every week, it will take her \$28 / \$4 = 7 more weeks to save \$28.
Therefore, it will take her 7 more weeks to have saved a total of \$60.
\#\#\#\# 7\\
\hline
+ Ext. reward & Jane has saved \$4 every week for 8 weeks, so she has saved a total of \$4 x 8 = \$32.
She wants to save a total of \$60, so she still needs to save \$60 - \$32 = \$28.
If she continues to save \$4 every week, it will take her \$28 / \$4 = 7 more weeks to save \$28.
Therefore, it will take her 7 more weeks to have saved a total of \$60.
\#\#\#\# 7\\
\hline
+ Ext + KL & Jane has saved \$4 every week for 8 weeks, so she has saved a total of \$4 x 8 = \$32.
She wants to save a total of \$60, so she still needs to save \$60 - \$32 = \$28.
If she saves \$4 every week, it will take her \$28 / \$4 = 7 more weeks to save \$28.
\#\#\#\# 7\\
\hline
+ CoVO & Jane has saved \$4 every week for the past 8 weeks, so she has saved a total of \$4 x 8 = \$32.
If she continues to save the same amount each week, we can set up an equation to find out how many more weeks it will take for her to have saved a total of \$60.
The equation is: \$32 + \$4x = \$60, where x represents the number of additional weeks.
To solve for x, we subtract \$32 from both sides of the equation: \$4x = \$60 - \$32.
Simplifying the right side of the equation, we have: \$4x = \$28.
To isolate x, we divide both sides of the equation by \$4: x = \$28 / \$4.
Simplifying the right side of the equation, we have: x = 7.
Therefore, it will take Jane 7 more weeks to have saved a total of \$60.
\#\#\#\# 7\\
\hline
+ CoVO + KL & Jane has saved \$4 every week for the past 8 weeks, so she has saved a total of \$4 x 8 = \$32.
If she continues to save the same amount each week, we need to find out how many more weeks it will take for her to have saved a total of \$60.
We can set up an equation to solve for the number of weeks.
\$32 + \$4 x (number of weeks) = \$60
Simplifying the equation, we have \$4 x (number of weeks) = \$60 - \$32
\$4 x (number of weeks) = \$28
Dividing both sides of the equation by \$4, we have (number of weeks) = \$28 / \$4
(number of weeks) = 7
Therefore, it will take Jane 7 more weeks to have saved a total of \$60.
\#\#\#\# 7\\
\hline
+ CoVO + Ext & Jane has saved \$4 every week for the past 8 weeks, so she has saved a total of \$4 x 8 = \$32.
If she continues to save \$4 each week, we need to find out how many more weeks it will take for her to have saved a total of \$60.
We can set up an equation to solve for the number of weeks:
\$32 + \$4 x (number of weeks) = \$60
Simplifying the equation, we have:
\$4 x (number of weeks) = \$60 - \$32
\$4 x (number of weeks) = \$28
Dividing both sides of the equation by \$4, we have:
(number of weeks) = \$28 / \$4
(number of weeks) = 7
Therefore, it will take Jane 7 more weeks to have saved a total of \$60.
\#\#\#\# 7\\
\hline
+ CoVO + Ext + KL & Jane has saved \$4 every week for the past 8 weeks, so she has saved a total of \$4 x 8 = \$32.
If she continues to save \$4 each week, we need to find out how many more weeks it will take for her to have saved a total of \$60.
We can set up an equation to solve for the number of weeks.
\$32 + \$4 x (number of weeks) = \$60
Simplifying the equation, we have \$4 x (number of weeks) = \$60 - \$32
\$4 x (number of weeks) = \$28
Dividing both sides of the equation by \$4, we have (number of weeks) = \$28 / \$4
(number of weeks) = 7
Therefore, it will take Jane 7 more weeks to have saved a total of \$60.
\#\#\#\# 7\\
\hline
\end{tabular}
}
\caption{\label{gsm8k_example_2} A qualitative example of GSM8K outputs for the base \texttt{MetaMath-Mistral-7} model and its tuned variants, where different models follow different solving strategies leading to the same results.}
\end{table*}

\begin{table*}
\centering
\resizebox{\textwidth}{!}{%
\begin{tabular}{|m{4.0cm}|m{5cm}|m{5cm}|}
\hline
\rowcolor{gray!20} Prompt & \multicolumn{2}{m{12cm}|}{Name a Spanish speaking country.}\\
\hline
& \multicolumn{2}{m{12cm}|}{[`Mexico.', `Colombia.', `Mexico.', `Spain is a Spanish-speaking country.', `Spain.', `Mexico is a Spanish-speaking country.', `Mexico.', `Colombia.', `Mexico is a Spanish-speaking country.', `Mexico.']}\\
\cline{2-3}
\multirow{-4}{*}{\texttt{Llama3.2-3B-Instruct}} & Utility: 5.98 & Distinct: 3 \\
\hline
& \multicolumn{2}{m{12cm}|}{[`Mexico', `One example is Mexico.', `Mexico', `Mexico is a Spanish-speaking country.', `Mexico.', `Spain is a country that is predominantly Spanish-speaking, with the primary official language being Castilian Spanish', `Colombia', `One Spanish speaking country is Chile', `Chile', `One of the Spanish-speaking countries is Spain, in europe']}\\
\cline{2-3}
\multirow{-6}{*}{+ CoVO} & Utility: 7.83 & Distinct: 7 \\
\hline
& \multicolumn{2}{m{12cm}|}{[`Mexico', `Mexico is an example of a Spanish-speaking country.', `Colombia', `Mexico', `Mexico is a Spanish-speaking country.', `Mexico is a Spanish speaking country.', `Colombia', `Colombia is a Spanish speaking country', 'Mexico.', `One Spanish-speaking country is Mexico.']}\\
\cline{2-3}
\multirow{-5}{*}{+ CoVO + KL } & Utility: 7.06 & Distinct: 5 \\
\hline
\rowcolor{gray!20} Prompt & \multicolumn{2}{m{12cm}|}{Pick a random valid US zip code.}\\
\hline
& \multicolumn{2}{m{12cm}|}{[`The random valid US zip code I picked is: 85351', `The random US zip code I picked is: 85353', `The random US zip code I picked is: 85401', `The random US zip code I picked is: 98382', `The random US zip code I picked is: 85451', `The random US zip code I picked is: 94043', `The random US zip code I picked is: 85323', `The random US zip code I picked is: 94043', `The random valid US zip code I picked is: 45429', `The random US zip code I picked is: 95814']}\\
\cline{2-3}
\multirow{-8}{*}{\texttt{Llama3.2-3B-Instruct}} & Utility: 1.79 & Distinct: 0 \\
\hline
& \multicolumn{2}{m{12cm}|}{[`Here is a randomly generated valid US zip code: 98104', `Beverly Hills, California 90210', `A random US zip code from my knowledge... It's 74114.', `Let's pick a random valid US zip code. We have our zip code: `32103`', `Here is a random valid US zip code: 92064', `The random US zip code I have chosen for you is: 85704', `I have randomly generated a valid US zip code for you: 94305 This is a valid zip code that serves the city of Cupertino, California.', `I picked: 98031', `The random US zip code is 85323', `I picked a valid US zip code. The zip code is 85032.']}\\
\cline{2-3}
\multirow{-9}{*}{+ CoVO} & Utility: 5.19 & Distinct: 8 \\
\hline
& \multicolumn{2}{m{12cm}|}{[`A random valid US zip code is: 85005', `The random US zip code I picked is 98524', `One random valid US zip code is: 95824', `The zip code is 98514', `The random US zip code I picked is: 92661', `The random US zip code I picked is: 84117', `The 97202 zip code, which is located in Portland, Oregon.', `The random US zip code I picked is 95404', `The selected US zip code is 97028', `One of my random picks for a valid US zip code is 48217.']}\\
\cline{2-3}
\multirow{-7}{*}{+ CoVO + KL } & Utility: 4.13 & Distinct: 4 \\
\hline
\end{tabular}
}
\caption{\label{llama_nb_examples1} Two qualitative examples of NoveltyBench outputs for the base \texttt{Llama3.2-3B-Instruct} model and its two tuned variants that show how our method enables a larger number of output templates.}
\end{table*}

\begin{table*}
\centering
\resizebox{\textwidth}{!}{%
\begin{tabular}{|m{4.0cm}|m{5cm}|m{5cm}|}
\hline
\rowcolor{gray!20} Prompt & \multicolumn{2}{m{12cm}|}{I am doing online shopping for a high-quality summer dress. What are the first 3 stores/websites I should check out? Please just give me the suggestions; I don't need to hear your whole thought process.}\\
\hline
& \multicolumn{2}{m{12cm}|}{[`1. Net-a-Porter 2. Moda Operandi 3. ASOS', `1. ASOS 2. Zara 3. Net-a-Porter', `1. Net-a-Porter 2. ASOS 3. Zara', `1. Net-a-Porter 2. ASOS 3. Zara', `1. Net-a-Porter 2. Moda Operandi 3. Zara', `1. Nordstrom 2. ASOS 3. Zara', `1. ASOS 2. Zara 3. Free People', `1. ASOS 2. ModCloth 3. Zara', `1. ASOS 2. Nordstrom 3. Zara', `1. Net-a-Porter 2. Zara 3. ASOS']}\\
\cline{2-3}
\multirow{-6}{*}{\texttt{Llama3.2-3B-Instruct}} & Utility: 2.82 & Distinct: 1 \\
\hline
& \multicolumn{2}{m{12cm}|}{[`ASOS, Zara, and Madewell', `Here are 3 stores you can check out for a high-quality summer dress:  1. Net-a-Porter 2. Zara 3. Madewell', `1. Zara.com  2. ModCloth.com  3. Net-A-Porter', `1. Nordstrom  2. Net-a-Porter 3. Zara', 'Zara, Madewell, and ASOS.', `1. ASOS 2. Madewell 3. Free People', `Here are three online stores/websites you can consider for a high-quality summer dress:  1. Zara 2. Madewell 3. Net-a-Porter', 'Here are 3 popular stores to start checking:  1. ASOS 2. Zara 3. Madewell', `Here are the top 3 stores to start:  1. ASOS 2. Zara 3. Nordstrom', [...]]}\\
\cline{2-3}
\multirow{-8}{*}{+ CoVO} & Utility: 6.40 & Distinct: 5 \\
\hline
& \multicolumn{2}{m{12cm}|}{[`Here are three online stores you should consider for high-quality summer dresses:  1. Net-a-Porter 2. ASOS 3. Moda Operandi', `1. Zara 2. Madewell 3. Nordstrom', `1. Nordstrom 2. ASOS 3. Net-a-Porter', `1. Nordstrom 2. Zara 3. Free People', `Here are the first 3 stores/websites to check out:  1. Net-a-Porter 2. Madewell 3. ASOS', `Zara, ASOS, and Nordstrom', `Here are three stores/websites you can start with:  1. ASOS 2. Net-a-Porter 3. Madewell', `Here are 3 popular stores/websites to check out for a high-quality summer dress:  1. Net-a-Porter 2. Zara 3. Madewell', `Here are three websites to consider:  1. ModCloth 2. ASOS 3. Nordstrom', `Here are the top 3 stores/websites to consider:  1. ASOS 2. Net-a-Porter 3. Nordstrom']}\\
\cline{2-3}
\multirow{-10}{*}{+ CoVO + KL } & Utility: 7.18 & Distinct: 5 \\
\hline
\rowcolor{gray!20} Prompt & \multicolumn{2}{m{12cm}|}{I am working on a memoir of a computer science PhD student who worked on machine translation in the 1990s. Suggest a single title and nothing else.}\\
\hline
& \multicolumn{2}{m{12cm}|}{[`"Bytes and Translations: A Journey Through the Dawn of Machine Translation"', `"Bridging Gaps: My Journey Through Machine Translation"', `"Code, Culture, and Translation"', `"Bits and Translations"', `"The Language Barrier"', `"Bits, Bytes, and Translation"', `"Bridging the Divide: A Life in Machine Translation"', `"Wires, Wheels, and Words: A Journey in Machine Translation"', `"The Language Barrier: A Memoir of Machine Learning\'s Early Days"', `"Bytes and Idioms: A Translation of Experience"']}\\
\cline{2-3}
\multirow{-8}{*}{\texttt{Llama3.2-3B-Instruct}} & Utility: 5.65 & Distinct: 5 \\
\hline
& \multicolumn{2}{m{12cm}|}{[`"Wires to Words":', `From the machine code', `"Bites of Code"', ``Bit by Byte: The Unlikely Translations of an Accidental Engineer'', `"Code for Life":', `"Bridging the Gap: Code, Culture, and the Search for Translation"', `"A Translation of Time"', `"Algorithms of Language"', `"Bits in Translation"', `The Machine that Almost Translated the World']}\\
\cline{2-3}
\multirow{-6}{*}{+ CoVO} & Utility: 4.18 & Distinct: 7 \\
\hline
& \multicolumn{2}{m{12cm}|}{[`"From Bits to Words"', `"Bytes and Borders"', `"Rebooting Language"', `"Translating My Mind: Adventures in Machine Learning's Early Years"', `"Bit by Bit Across the Divide"', `"Rebooting Eden: A Memoir of Early Days in Machine Translation"', `Algorithms for a New Language', `"Translating the Past: My Journey in the Dawn of Machine Translation"', `"A Language to Translate"', `"Coding the Universe, One Translation at a Time"']}\\
\cline{2-3}
\multirow{-7}{*}{+ CoVO + KL } & Utility: 7.15 & Distinct: 7 \\
\hline
\end{tabular}
}
\caption{\label{llama_nb_examples2} Two qualitative examples of NoveltyBench outputs for the base \texttt{Llama3.2-3B-Instruct} model and its two tuned variants that show how our method enables a diverse, semantic set of outputs.}
\end{table*}

\begin{table*}
\centering
\resizebox{\textwidth}{!}{%
\begin{tabular}{|m{4.0cm}|m{5cm}|m{5cm}|}
\hline
\rowcolor{gray!20} Prompt & \multicolumn{2}{m{12cm}|}{Pretend to pick a card from the top of a standard deck of cards. What card did you pick?}\\
\hline
& \multicolumn{2}{m{12cm}|}{[`I picked the Ace of Spades from the top of a standard deck of cards.', `I picked the Ace of Spades from the top of a standard deck of cards.', `I picked the Ace of Spades from the top of a standard deck of cards.', `I picked the Ace of Spades from the top of a standard deck of cards!', `I picked the Ace of Spades from the top of the deck!', `I picked the Ace of Spades from the top of a standard deck of cards!', `I picked the Ace of Spades from the top of a standard deck of cards!', `I picked the Ace of Spades from the top of a standard deck of cards!', `I picked the Ace of Spades from the top of a standard deck of cards.', `I picked the Ace of Spades from the top of the standard deck of 52 playing cards!']}\\
\cline{2-3}
\multirow{-10}{*}{\texttt{Qwen2.5-7B-Instruct}} & Utility: 2.02 & Distinct: 0 \\
\hline
& \multicolumn{2}{m{12cm}|}{[`I picked the Ace of Spades from the top of a standard deck of 52 playing cards.', `I picked the Ace of Spades from the top of a standard deck of cards. This is a common choice often seen in card tricks and magic shows due to its prominent position at the top of the deck.', `I picked the Ace of Spades from the top of a standard deck of cards.', `Since I'm a digital AI assistant and don't have physical access to a deck of cards, I'll just pick a card from a standard 52-card deck. Let's say I picked the Ace of Spades. Remember, this is just a digital simulation!", `I've picked the Jack of Hearts from the top of a standard deck of cards. Would you like to know more about this card or play a game with it?', `I've picked the Ace of Spades from the top of a standard deck of cards. It's a classic choice often used in magic tricks and card games due to its simplicity and prominence on most decks.", `I picked the Ace of Spades from the top of a standard deck of cards.', [...]
]}\\
\cline{2-3}
\multirow{-14}{*}{+ CoVO} & Utility: 2.84 & Distinct: 1 \\
\hline
& \multicolumn{2}{m{12cm}|}{[`I've picked the top card from a standard deck of 52 playing cards. Since I can't actually see or touch physical cards, I'll pretend it's a very special and magical draw! Let's say I've picked the Ace of Spades. Enjoy your imaginary card game!', `I picked the Ace of Spades from the top of a standard deck of cards.', `Ah, let's see... I'll pretend to pick a card from the top of a standard deck. Ah, how about the Queen of Hearts? Enjoy the game!', `I picked the Ace of Spades from the top of a standard deck of 52 playing cards.', `Ah, I see you're asking for a bit of fun! Since I'm an AI and don't have the ability to physically pick a card from a deck, let's just pretend. If I were to pick a card from the top of a standard deck, it could be any one of the 52 cards. For the sake of this game, let's say the card I picked is the Ace of Spades. Would you like me to tell you about the Ace of Spades or do something with it in our imaginary game?', `I picked the Ace of Spades from the top of a standard deck of cards. Would you like to play a game or do something interesting with this card?', `I picked the Ace of Spades from the top of a standard deck of cards.', `I've picked the Jack of Hearts from the top of a standard deck of 52 playing cards. Enjoy your game!', `I've picked the Jack of Hearts from the top of a standard deck of cards. Would you like to play a game or do something with this card?', `I picked the Ace of Spades from the top of a standard deck of cards. While I don't have the ability to physically manipulate objects, I can certainly imagine or simulate picking this card for our conversation!']}\\
\cline{2-3}
\multirow{-21}{*}{+ CoVO + KL} & Utility: 3.31 & Distinct: 1 \\
\hline
\end{tabular}
}
\caption{\label{qwen_nb_example_1} A qualitative example of NoveltyBench outputs for the base \texttt{Qwen2.5-7B-Instruct} model and its two tuned variants that shows how our methods allow for greater content variation, despite adopting the same template.}
\end{table*}

\begin{table*}
\centering
\resizebox{\textwidth}{!}{%
\begin{tabular}{|m{4.0cm}|m{5cm}|m{5cm}|}
\hline
\rowcolor{gray!20} Prompt & \multicolumn{2}{m{12cm}|}{Generate the safest first digit for my 12 digit password? Give me exactly one digit.}\\
\hline
& \multicolumn{2}{m{12cm}|}{[`1', `1', `1', `1', `1', `1', `1', `1', `1', `1']}\\
\cline{2-3}
\multirow{-2}{*}{\texttt{Qwen2.5-7B-Instruct}} & Utility: 0.45 & Distinct: 0 \\
\hline
& \multicolumn{2}{m{12cm}|}{[`To ensure your 12-digit password is as secure as possible, it's best to use a combination of letters (both uppercase and lowercase), numbers, and special characters. However, since you're asking for a single digit, I'll provide a recommendation for the safest first digit. The safest and most commonly used digits in a password are not those that are too predictable or common. Therefore, avoiding 0, 1, 2, 3, 4, 5, 6, 7, 8, and 9 based on predictability isn't practical. Instead, we can choose a digit that is less frequently used at the beginning of passwords. Given this context, a good choice would be **9**. It's relatively uncommon as the first digit in a sequence and can add an extra layer of security to your password. So, the safest first digit for your 12-digit password is **9**.', `1', `For creating a secure password, it's important to consider not just the first digit but the entire composition of the password. However, if you need a single digit as the safest first digit for your 12-digit password, choosing a number that is less common or harder to guess can be beneficial. Given the standard range of digits (0-9), and assuming no additional context about potential attackers' guesses, a good choice could be: **7** This digit is relatively uncommon in simple or easily guessed passwords and adds a layer of security. Remember, a strong password should also include a mix of uppercase and lowercase letters, numbers, and special characters, and it should not be based on personal information.', [...]
]}\\
\cline{2-3}
\multirow{-20}{*}{+ CoVO} & Utility: 1.18 & Distinct: 6 \\
\hline
& \multicolumn{2}{m{12cm}|}{[`The safest first digit for your 12-digit password would be `0`. However, it's important to note that while `0` is a valid and safe starting digit, the overall security of your password depends on the combination of all digits and any other characters or factors you might use. Ensure the rest of your password includes a mix of uppercase and lowercase letters, numbers, and special characters to enhance its security.", `Certainly! For the sake of security, you should choose a digit that is not commonly used as the first digit in passwords, such as "0" or "1", which are often chosen due to their simplicity. A safer choice would be: **3** This helps reduce the likelihood of your password being guessed or cracked based on common patterns.', `The safest first digit for your 12-digit password would be **9**. Using a less common digit as the first character can add an extra layer of security to your password.', `1', `1', `To ensure your 12-digit password is as secure as possible, it's important to use a digit that reduces the risk of brute-force attacks. The safest first digit would be one that is not easily guessable and adds complexity to your password. Given these considerations, I recommend starting with: **8** This digit is commonly used but still provides a good level of security when combined with other characters in a random or complex manner.", [...]
]}\\
\cline{2-3}
\multirow{-20}{*}{+ CoVO + KL } & Utility: 1.13 & Distinct: 6 \\
\hline
\end{tabular}
}
\caption{\label{qwen_nb_example_2} A qualitative example of NoveltyBench outputs for the base \texttt{Qwen2.5-7B-Instruct} model and its two tuned variants that shows how our methods allow for greater variation, also in terms of output templates.}
\end{table*}

\end{document}